\documentclass[%
 reprint,
superscriptaddress,
 amsmath,amssymb,
 aps,
 pre,
]{revtex4-1}

\usepackage{graphicx}
\usepackage{dcolumn}
\usepackage{bm}
\usepackage{xcolor}
\usepackage[bf]{subfigure}
\usepackage{hyperref}

\begin{document}

\preprint{APS/123-QED}

\title{A keyword-driven approach to science}

\author{Henrique Ferraz de Arruda}
\author{Luciano da Fontoura Costa}
\affiliation{
 S\~ao Carlos Institute of Physics,
University of S\~ao Paulo, PO Box 369,
13560-970, S\~ao Carlos, SP, Brazil
}

\date{\today}

\begin{abstract}
To a good extent, words can be understood as corresponding to patterns or categories that
appeared in order to represent concepts and structures that are particularly important or
useful in a given time and space.  Words are characterized by not being completely general
nor specific, in the sense that the same word can be instantiated or related to several
different contexts, depending on specific situations.  Indeed, the way in which words are
instantiated and associated represents a particularly interesting aspect that can substantially help to better understand the context in which they are employed.  Scientific words are no  exception to that.  In the present work, we approach the associations between a set of particularly relevant words in the sense of being not only frequently used in several areas, but also representing concepts that are currently related to some of the main standing challenges in science.   More specifically, the study reported here takes into account the words ``prediction'', ``model'', ``optimization'', ``complex'', ``entropy'', ``random'', ``deterministic'', ``pattern'', and ``database''. In order to complement the analysis, we also obtain a network representing the relationship between the adopted areas.  Many interesting results were found. First and foremost, several of the words were observed to have markedly distinct associations in different areas. Biology was found to be related to computer science, sharing associations with databases. Furthermore, for most of the cases,  the words ``complex'', ``model'',  and ``prediction'' were observed to have several strong associations.
\end{abstract}

\maketitle


\section{\label{sec:introduction}Introduction}

Since its origins centuries ago as a mostly unified endeavor, science has subsequently branched into a complex hierarchy involving several areas and
sub-areas~\cite{rosvall2011multilevel}, as a consequence of inevitable progressive specialization required to tackle the growing complexity of several natural world phenomena.  

Currently, there is a large number of areas in science and technology~\cite{morillo2003interdisciplinarity,small2006tracking}, each with its
respective focus, which also impacts several other respective characteristics such as the type of approach (e.g.~statistic, theoretical, numerical, etc.),  employed data (e.g.~large databases, simulated, continuous/discrete, etc.),  respective dissemination (e.g.~books, articles, reports, etc.) and, last but not least, the use of specific concepts and words~\cite{whittaker1989creativity}.

Indeed, as a consequence of the intrinsic differences between the several scientific areas, even the use of more general words and their associations with other
concepts can be expected to vary.  For instance, take the word \emph{complex}.  In areas related to physical science, this concept will likely be more frequently associated with concepts such as information, entropy, and computational cost.  On the other hand, the same word \emph{complexity} may relate more frequently with words such as
brain, immune system, and animal development in a more biological context.

The intrinsic variation in the use of same words in different scientific and technological areas therefore defines a particularly interesting
problem on itself, with potential for contributing not only to a better comprehension of how words are typically used in distinct areas, but also with potential to achieving a more comprehensive understanding of  science itself.  Several are also the possible more practical applications of a better knowledge about the contextual use of words, such as in devising improved approaches to translation, automated semantic analysis, summarization, among many other interesting possibilities.

An additional result allowed by the study of the specific use of words in different scientific contexts regards the identification of words that
are used in more or less uniformly manner within a specific set of  scientific areas.  For instance, a word that is employed quite differently in diverse areas is more likely to be misunderstood.   This type of information also has many other possible interesting applications, such as in identifying words that need to be presented with special care by a given area when trying to address another specific area, or even a multidisciplinary set of contexts.

The concept of co-occurrence of words is employed in many applications, such as understanding how humans use the words~\cite{sicilia2003extension,cavnar1994n} and text classification~\cite{tripathy2016classification,khreisat2006arabic,peng2003combining}. By considering  standard text classification methods, the counts of sets of consecutive words, called $n$-grams, are frequently employed~\cite{khreisat2009machine,al2008improving, zampieri2013using}.  Observe that the concept of co-occurrence in these works related to the adjacency of two or more words along the text, while in the present work we take into account the presence of two or more words in a single entry of Wikipedia as a more general indication of their possible relationship.   

In approaches based in co-occurrence, $n$-grams have also been employed to assign labels to scientific papers organized into groups according to citations~\cite{silva2016using, ceribeli2021coupled,benatti2021enriching}.  Furthermore, many studies have dealt with co-occurrence of words by using networks~\cite{brede2008patterns}, typically in studies comparing between different writing styles~\cite{amancio2012identification,de2016using,marinho2016authorship} (e.g., analysis of literary movements~\cite{amancio2012identification} or different types of text~\cite{de2016using}, authorship attribution~\cite{marinho2016authorship}, among others). More recently, techniques related to embedding~\cite{hu2016different,pennington2014glove,mikolov2013efficient} have considered the co-occurrence of words (e.g., \emph{glove}~\cite{pennington2014glove} and \emph{word2vec}~\cite{mikolov2013efficient}).

The present work aims at investigating in a quantitative, objective manner the uniformity/heterogeneity of the use of specific scientific and technological terms.  Instead of considering a large number of technical words, which would constitute an interesting related research on itself, we focus on a smaller set of terms that represent some of the main tendencies in science and technology~\cite{costa2020learning}, namely: ``complex'', ``database'', ``deterministic'', ``entropy'', ``model'', ``optimization'', ``pattern'', ``prediction'', and ``random''.  Some other interesting words, such as \emph{causal} could not be incorporated because of the relatively small number of appearances in the adopted database.  Such an approach, focused on target words, has some interesting advantages while still allowing a good illustration of the prospect of identifying the usage uniformity.  First and foremost, we have that the results can  be discussed individually, therefore allowing a more substantive and comprehensive analysis than could be achieved if dozens or hundreds of terms had been used instead.  Then, we have that there are not so many words that are general enough to be found in several areas and still have plural interpretations.

We start our analysis by considering only the frequency of words in the texts. In order to understand the relationship between words, we considered the co-occurrence of words. As the next step, we modeled the word relationships as networks, which are henceforth called \emph{colocalization} networks. From these networks, we analyzed both the pairwise relationships as well as the groups of words. Finally, aiming at understanding the relationship among the areas in terms of these networks, we propose a method for summarizing the colocalization into a single network, namely \emph{summary} network. Many interesting results were obtained. For example, the word ``database'' was found to play an essential role in the area of \emph{biology}.  In addition, the word ``optimization'' was identified to play a central role in the area of \emph{computer science} and \emph{mathematics}, but with markedly different usages. Furthermore, the results corroborated that the words can be used in different manners depending on the study area. 

The remainder of this paper is organized as follows. Section~\ref{sec:words} discusses the choice of the employed scientific terms and fields of knowledge. In the same section, the dataset and the network model used in the analysis are described. In Section~\ref{sec:results}, we present the obtained results as well as their discussion. Finally, in Section~\ref{sec:conc}, we present the conclusions and perspectives of future works.

\section{\label{sec:materials}Materials and Methods}
In this section, we discuss the choice of the analyzed words and areas. Next, the dataset as well as the employed methodology for data modeling and analysis are described.

\subsection{The Selected Words}
\label{sec:words}

In the following we provide a brief presentation of the justification and motivation for
the choice of each \emph{scientific word}, as well as some discussion about their 
characteristics and relationships with other terms.  The main reference for the choice
of this terms, in addition to their frequent modern usage, stems from the key  activities
of predicting and modeling, which forms the basis of the scientific approach.

\emph{\underline{Prediction}}: This corresponds to the main motivation in science and
the scientific method.  In other words, the potential of each developed model ultimately
related to is ability to \emph{predict} phenomena in the physical world.  Therefore,
prediction could be expected to related, in more general terms, with terms such as
model, complexity, randomness, and database.  However, these possible associations can
be expected to vary from area to area.  For instance, prediction could be expected to
play a more central role in areas directly related to the physical world, as physics,
neuroscience (e.g, \cite{brown2012role}) and biology (e.g., \cite{stepanchikova2003prediction}).

\emph{\underline{Model}}: Modeling is the principal manner through which scientists
approach natural phenomena in order to better understand them and make respective
predictions.  Though central in science and technology, the relationships of this
term are expected to vary in different fields, as a consequence of the types of
models and modeling approaches.  For instance, modeling approaches relying on 
large quantities of data, as typically observed in areas such as neuroscience and 
biology, are more likely to be related to terms as databases~\cite{aderem2005systems}.  Contrariwise, modeling 
approaches in areas as physics and mathematics may be more closely related to 
concepts such as optimization (e.g.,\cite{najafi2020physics}).

\emph{\underline{Optimization}}: The development of a scientific model is intrinsically
related to the problem of optimization.  Indeed, models are made progressively more
accurate by considering the quality of the respective predictions, which is therefore
optimized.  The development of effective optimization methods, be them of computational
or more general nature, represents a key aspect in science and technology given
its potential to help solving several current challenging problems.
The role of the term optimization in different areas is likely to vary
as a consequence of the types of respective data analysis and modeling approaches.

\emph{\underline{Complex}}~\cite{weaver1991science}: One of the greatest scientific challenges corresponds to 
complexity.  Though not precisely defined, complexity subsumes many multiple component
systems undergoing non-linear dynamics.  These problems are challenging not only
regarding their theoretical modeling, but also respective attempts of computational
analysis and simulations.  As such, complexity becomes inherently linked to 
modeling, prediction, entropy, databases, etc.  It is expected that the meaning and
relationships between complexity and other scientific terms will vary respectively to
distinct areas.  The importance of complexity is directly reflected in the onset of
new areas such as \emph{complex systems}~\cite{weaver1991science} and \emph{complex networks}~\cite{costa2007characterization}.

\emph{\underline{Entropy}}~\cite{cover1999elements}: Developed independently in statistical physics and
information theory, the concept of entropy has attained a central role in science
and technology thanks not only to its powerful conceptualization, but also to the
many applications that have been respectively motivated.  In particular, entropy has
been considered as a possible quantification of complexity (e.g.,~\cite{pincus1991approximate}).  Given its stochastic
nature, entropy is also intrinsically related to randomness.  This term is expected
to have varying applications and relationships in distinct areas.  For instance, as
already observed, in physics it will be more directly related to statistical mechanics,
being more associated with information in computer science.

\emph{\underline{Random}}: Most real-world phenomena involves, to some extent, 
some randomness, may it be associated to measurement accuracy or experimental error.
As such, randomness constitutes an important concept in every applied science.
The application of this concept in different areas will reflect the accuracy of the
experimental data, as well as the types of models adopted for respective explanation.

\emph{\underline{Deterministic}}: Deterministic opposes randomness.  It remains a
challenging basic question whether nature is deterministic or not. This terms can
be expected to generally relate with concepts such as randomness (by opposition), 
modeling, database, etc.

\emph{\underline{Pattern}}: Pattern is a generic name given to basic entities in
modeling, corresponding possibly to a signal, an image, etc~\cite{fukunaga2013introduction}.  Reflecting the human
tendency to associate names to most entities, the problem of pattern recognition
plays a central role in science and technology, as it is ultimately necessary for
the automatizing of most human activities.  Generally speaking, the term pattern
is therefore expected to relate to random, complex, database, among others, but
it is likely to assume specific relationships in distinct areas.

\emph{\underline{Database}}: Given that scientific modeling relies critically on
the comparison of predictions/simulations with real-world quantitative measurements,
the importance of the concept of databases has progressively increased, giving rise
to research areas such as \emph{database}~\cite{mullins2002database}, \emph{data science}~\cite{dhar2013data}, and \emph{eScience}~\cite{hey2009jim}.  
Basically, a database corresponds to a logically organized dataset with some minimal
quality.  This term is potentially related to areas such as pattern, complex and model, 
but its associations can be expected to vary from an area to another.

\subsection{The Selected Fields} 
In order to investigate how the associations between the considered scientific
terms vary within distinct fields, it is necessary to select a representative group
of these areas.  The areas considered in the present work were chosen because, by 
their own nature, they are likely to incorporate a substantial number of entries
related to the selected scientific terms.  Thus, we have general areas as \emph{biology},
\emph{mathematics}, \emph{computer science} and \emph{physics}.  In addition, we also adopted some
relatively more specific areas --- more specifically \emph{dynamical systems} and
\emph{neuroscience} --- because of their potential relationship with the considered
scientific terms.

\subsection{The Dataset}
In order to select texts for the considered fields, we considered the \emph{Wikipedia}~\footnote{\url{https://www.wikipedia.org/}}, which is a free online encyclopedia created from contributors that maintain the content update. This platform is multilingual, but here we considered only the English version. For each described subject, corresponding to a respective page, there are hyperlinks connecting to other pages.  Some subjects are organized into subcategories. Here, we considered the pages cited from the articles listed in the following categories:

\begin{itemize}
   \item \emph{Physics}: Category:Concepts in physics; 
   \item \emph{Mathematics}: Category:Mathematics;
   \item \emph{Computer science}: Category:Computer science;
   \item \emph{Dynamical systems}: Category:Dynamical systems;
   \item \emph{Biology}: Category:Biology;
   \item \emph{Neuroscience}: Category:Neuroscience.
\end{itemize}

Furthermore, all the scientific words shown in Section~\ref{sec:words}, are considered in this work. Because some words are similar but not exactly the same in the text, before creating the network, the following pairs of words were considered as being the same: ``complex'' and ``complexity'', ``model'' and ``modeling'', ``prediction'' and ``predict'', and ``optimization'' and ``optimize''.

\subsection{Words colocalization network}
\label{sec:words_net}
We considered a network of co-occurrence of words in texts, which we call \emph{words colocalization network}. Observe that this is different from the standard co-occurrence networks because here the entire documents are considered, and not only the immediate adjacency between words.  In other words, two of the considered terms are understood to localize whenever they both occur in a same text.  Before creating this network, the texts are pre-processed. First, the part of speech (POS) tags, e.g.~nouns and verbs, is identified for all the words~\cite{ratnaparkhi1996maximum}.
By considering the identified POS tags, we lemmatize all the tokens using Wordnet~\cite{sigman2002global,miller1990introduction}. More specifically, lemmatization consists of the simplification of some words by their respective lemma. For example, the different verb tenses conjugation ``study'' ``studying'' ``studied'' are represented by the same lemma ``study''. 

The colocalization networks are created by considering all files for each field, where each word represents a node, and the edges are assigned when two words co-occur in at least one file. The weights are computed as the number of files leading to co-occurrence. As with the previous models, the resultant network is weighted and undirected.

\subsection{Summary network}
\label{sec:summary_net}
In order to interpret the relationship between the networks proposed in Section~\ref{sec:words_net}, we visualize relationships between the networks, $g_1,g_2, \dots, g_n$. More specifically, in a summarized version each node represents a network ($g_i$), while the edges are computed according to differences between their respective adjacency matrix. First, the weighted networks are converted to unweighted, according to a given threshold. All the $g_i$ edge weights are normalized by their highest value. The edges with weight lower than a threshold, $\tau$, are then removed. Here, for the considered summarized networks, we empirically define $\tau=0.5$. 
Edges are assigned to all possible connections between nodes $i$ to $j$ of the summarized network. Their weights are calculated as the number of the edges of $g_i$ that also take part in $g_j$.

\section{\label{sec:results} Results and discussions}
This section starts our analysis by considering word counts statistics and the co-occurrences of words in documents. Next, the word and field networks are considered. For all of the analyzed information, we discuss possible relationships between the scientific fields and words.

\subsection{Word statistics}
\begin{figure}[!ht]
  \centering
    \includegraphics[width=0.49\textwidth]{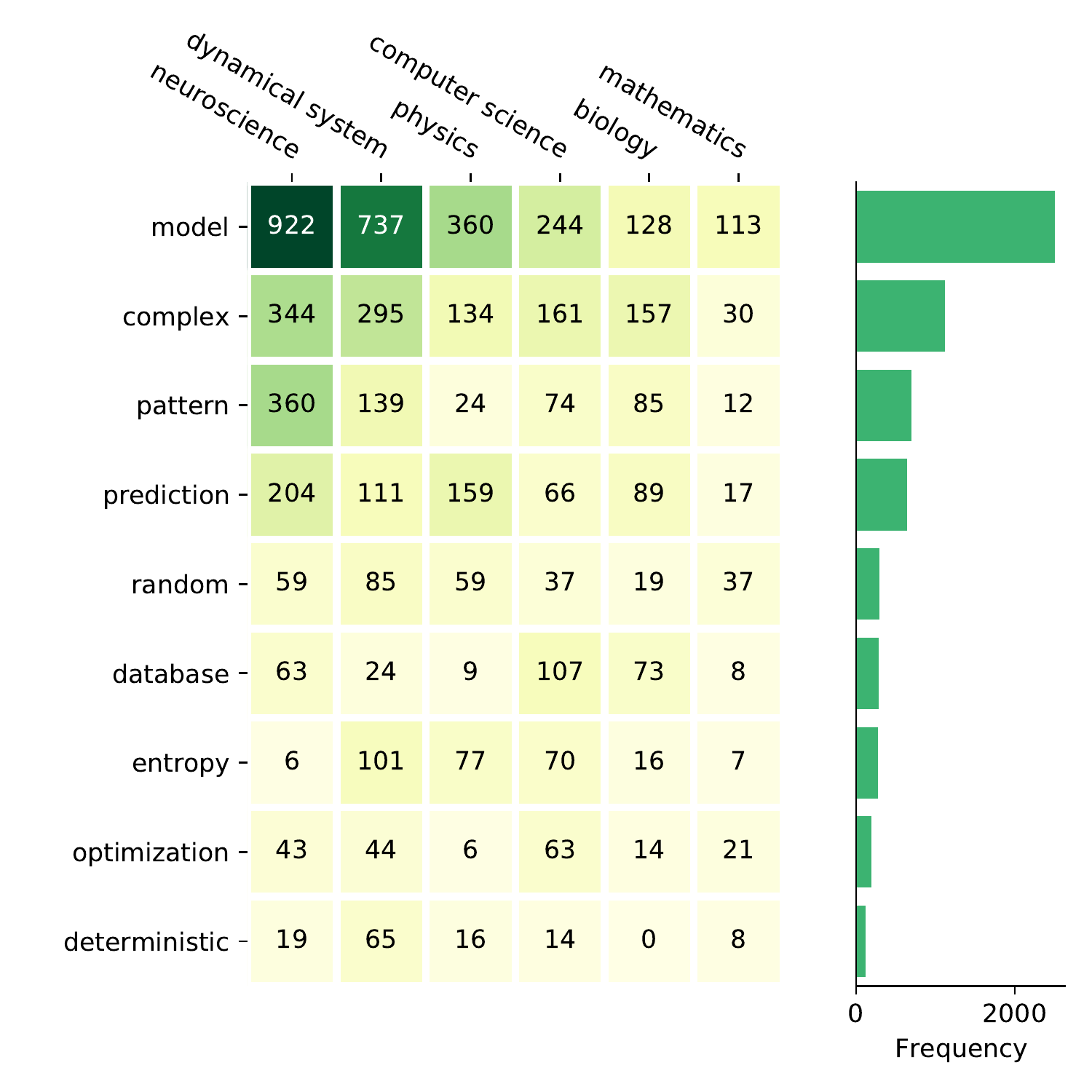}
  \caption {Number of the chosen words for each of the considered areas, as well as their respective total considering all areas.}
  \label{fig:res_comparison_counts}
\end{figure}

We start our analysis by considering the frequencies of words in each of the considered areas (see Figure~\ref{fig:res_comparison_counts}). This figure depicts the absolute value of each word found in each area. So, in general, the areas with bigger texts tend to have the highest frequencies for almost all words. The most frequent word in all areas is \emph{model}, which evidences its particular importance in science.  The second most frequent word is \emph{complex}, followed by \emph{pattern} and \emph{prediction}. In contrast with these results, \emph{deterministic} resulted in the least frequently used word.

\begin{figure*}[!ht]
  \centering
    \subfigure[\ complex]{\includegraphics[width=0.325\textwidth]{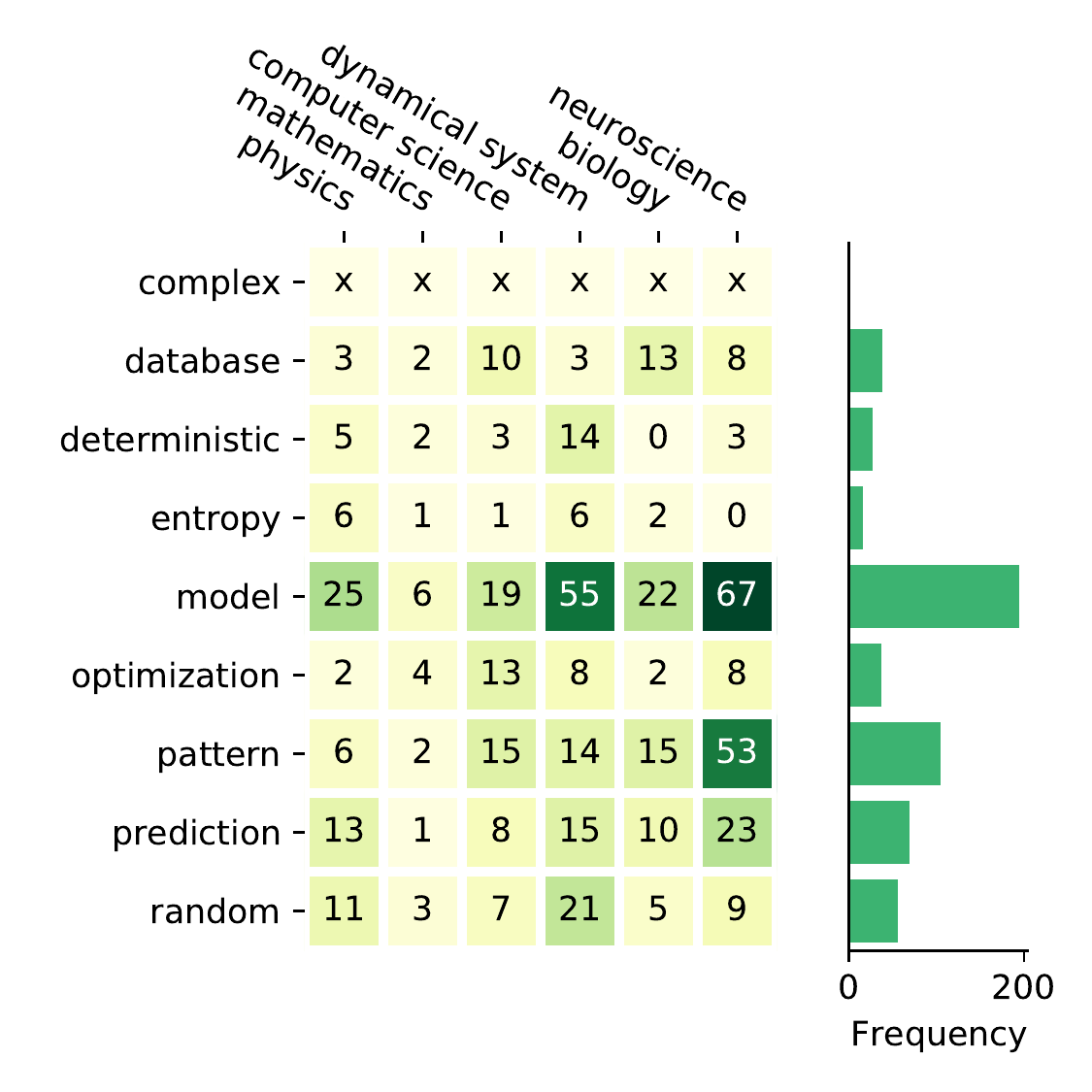}}
    \subfigure[\ database]{\includegraphics[width=0.325\textwidth]{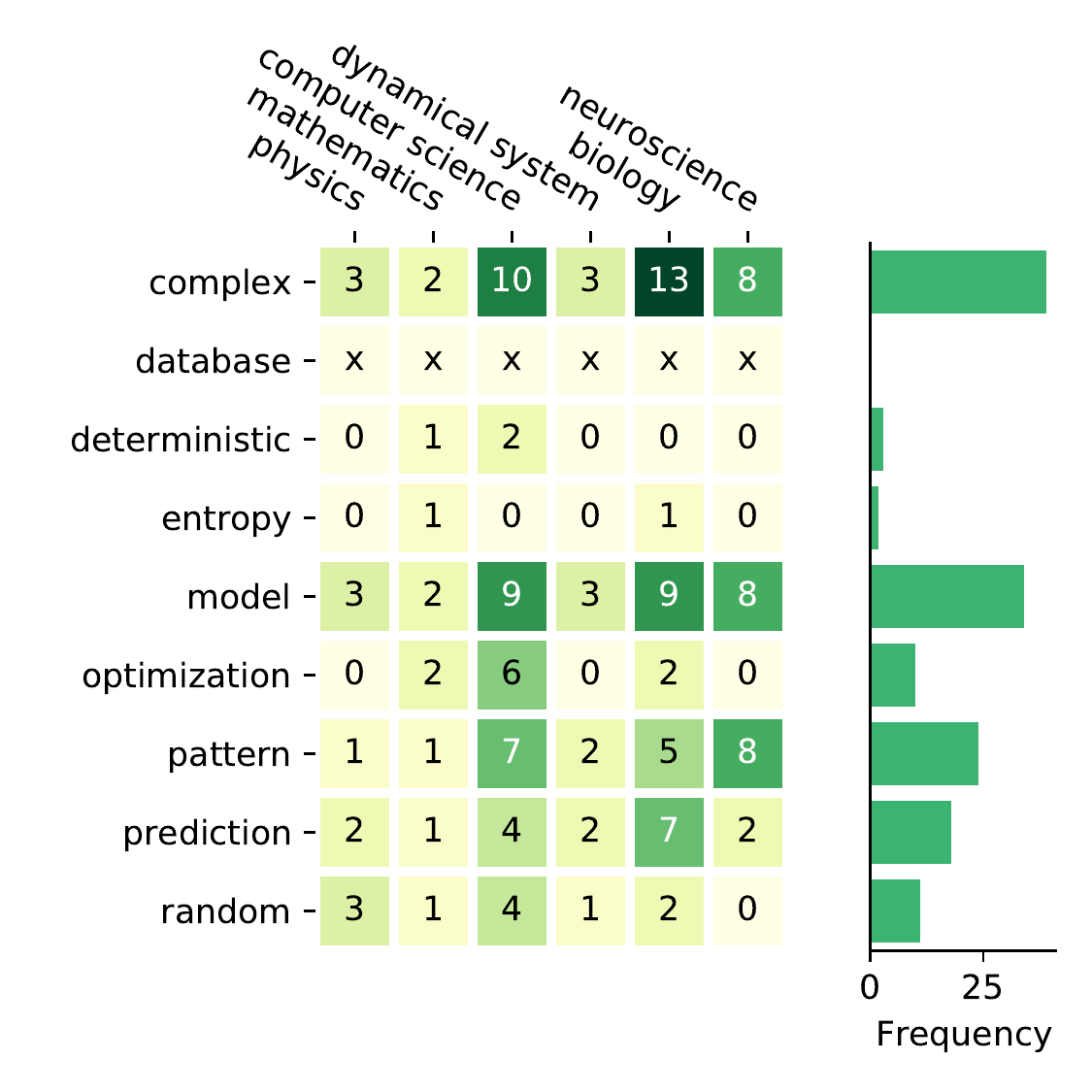}}
    \subfigure[\ deterministic]{\includegraphics[width=0.325\textwidth]{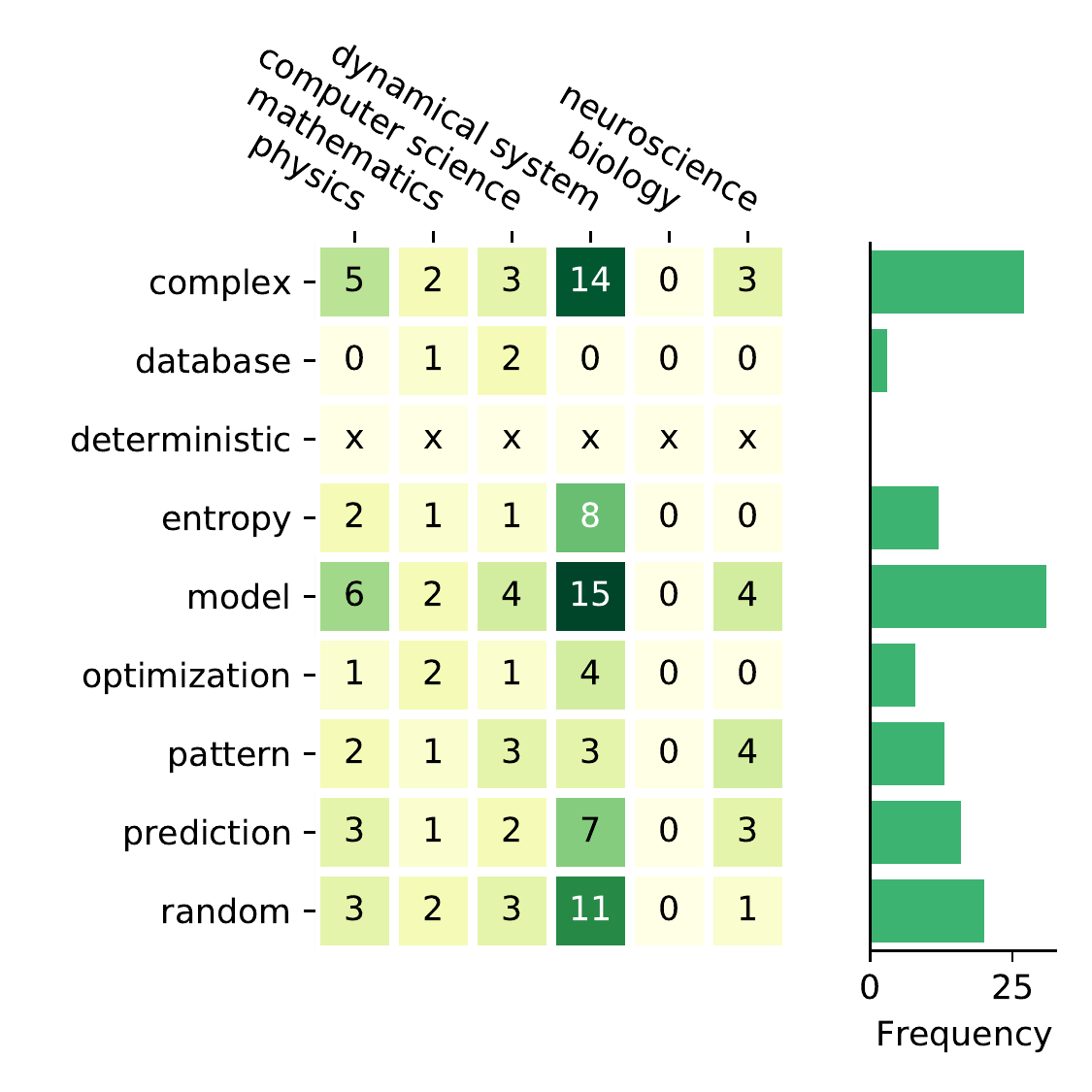}}
    
    \ 
    
    \subfigure[\ entropy]{\includegraphics[width=0.325\textwidth]{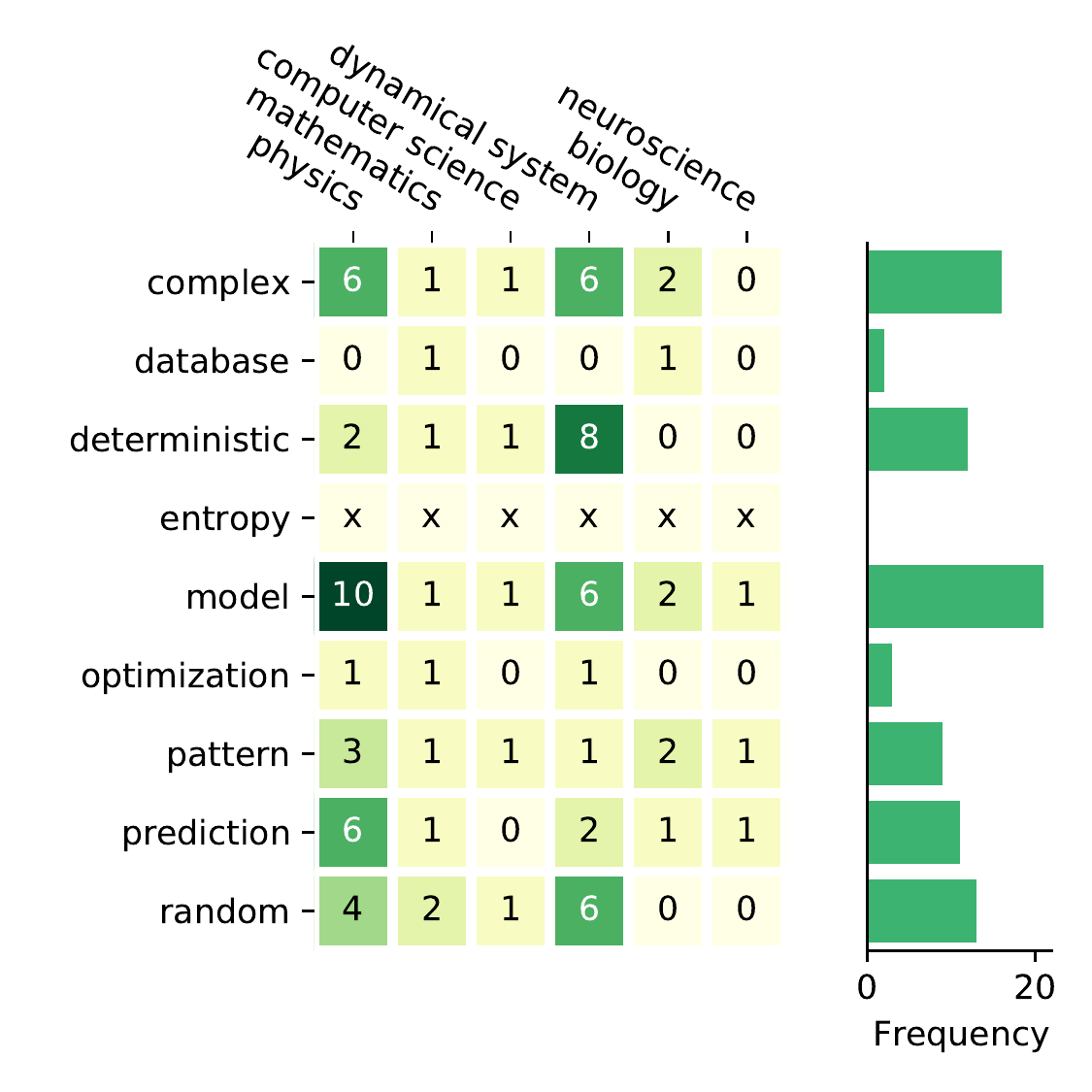}}
    \subfigure[\ model]{\includegraphics[width=0.325\textwidth]{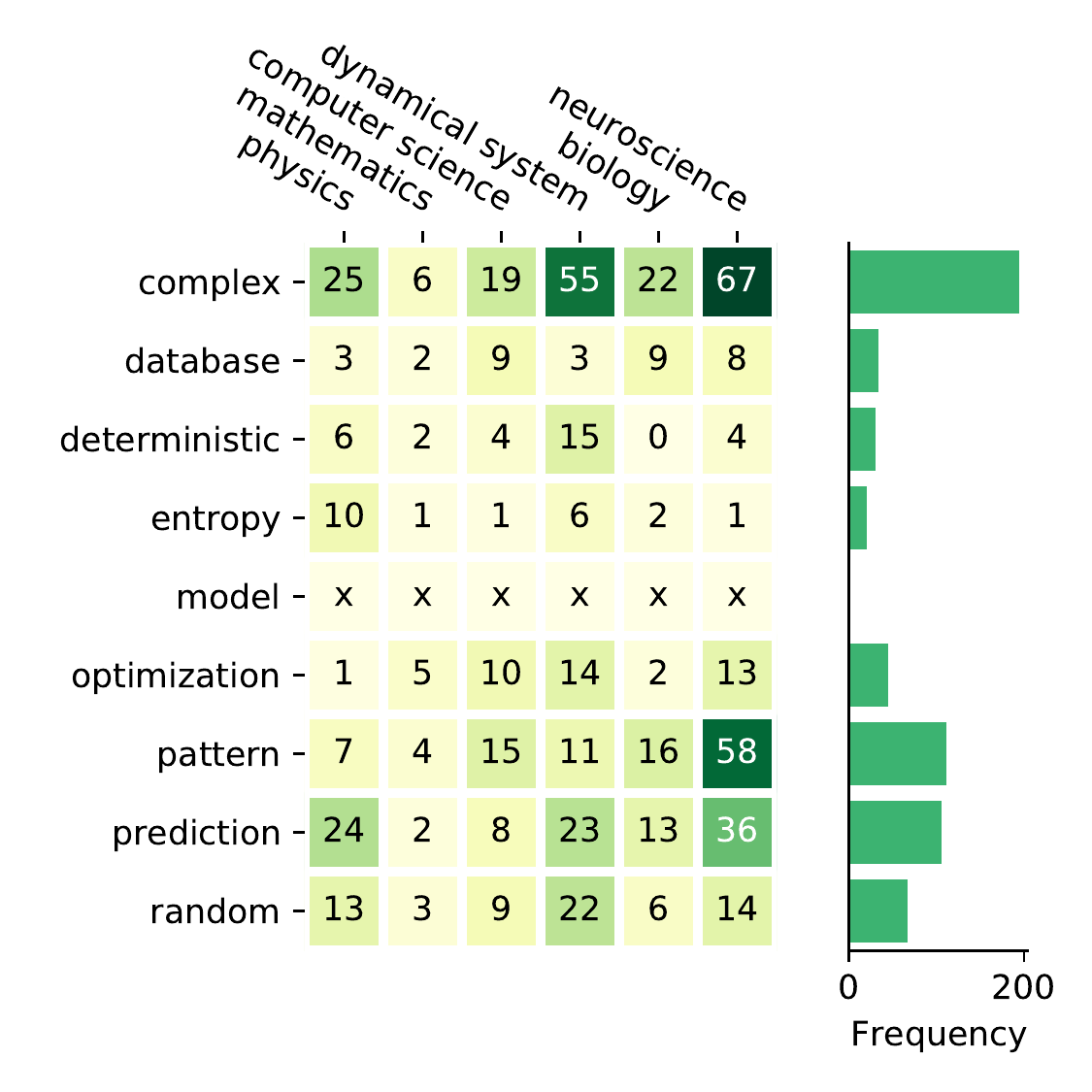}}
    \subfigure[\ optimization]{\includegraphics[width=0.325\textwidth]{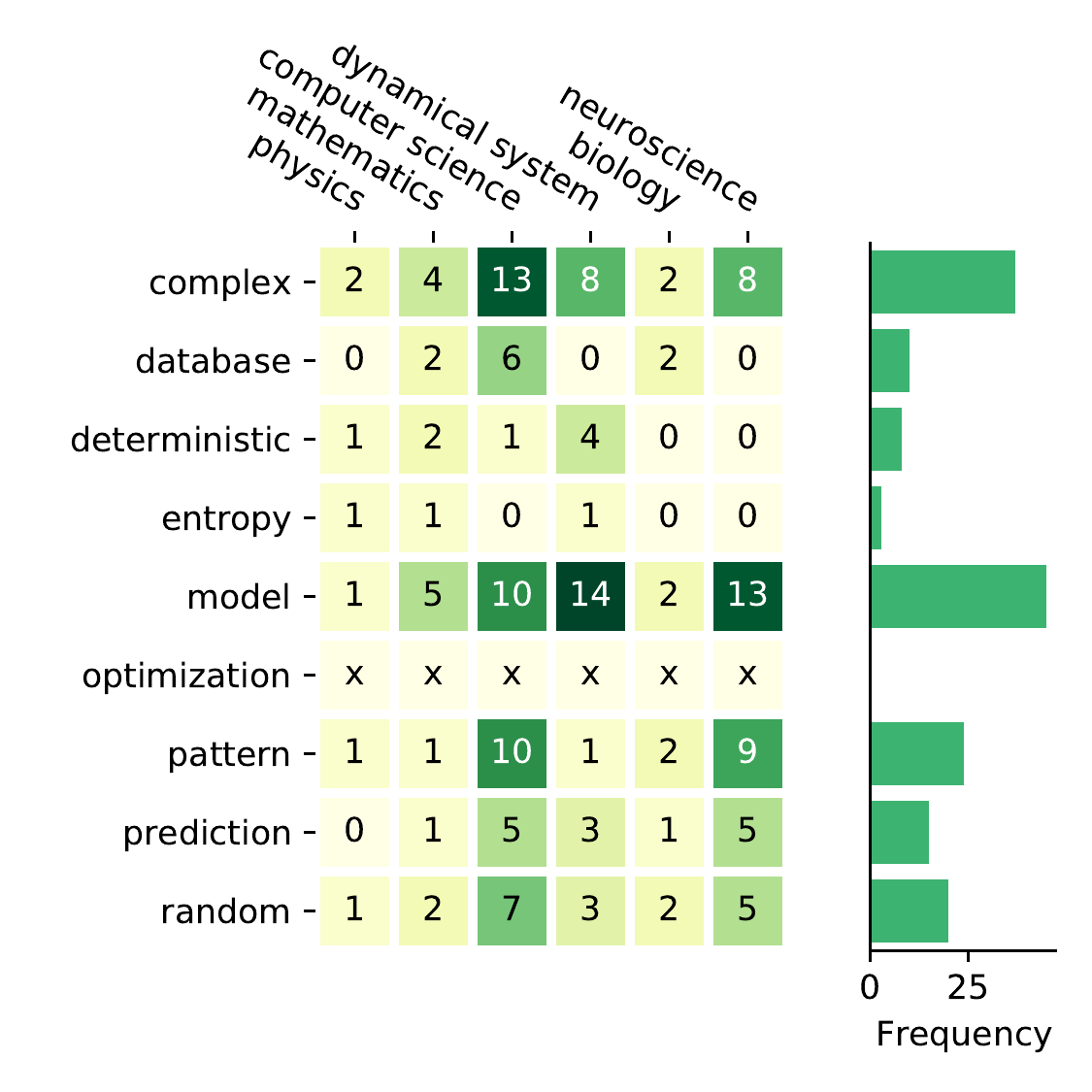}}
    
    \ 
    
    \subfigure[\ pattern]{\includegraphics[width=0.325\textwidth]{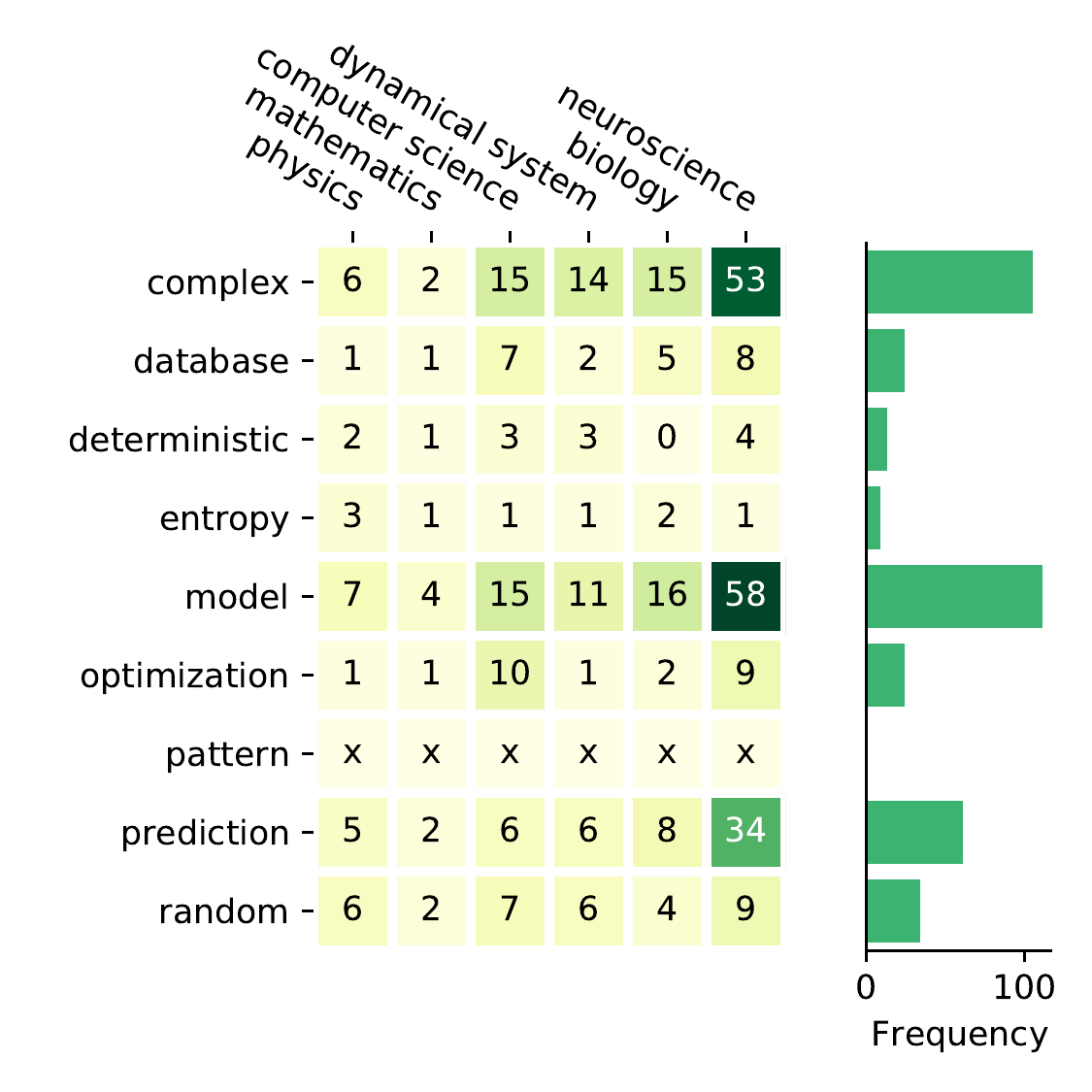}}
    \subfigure[\ prediction]{\includegraphics[width=0.325\textwidth]{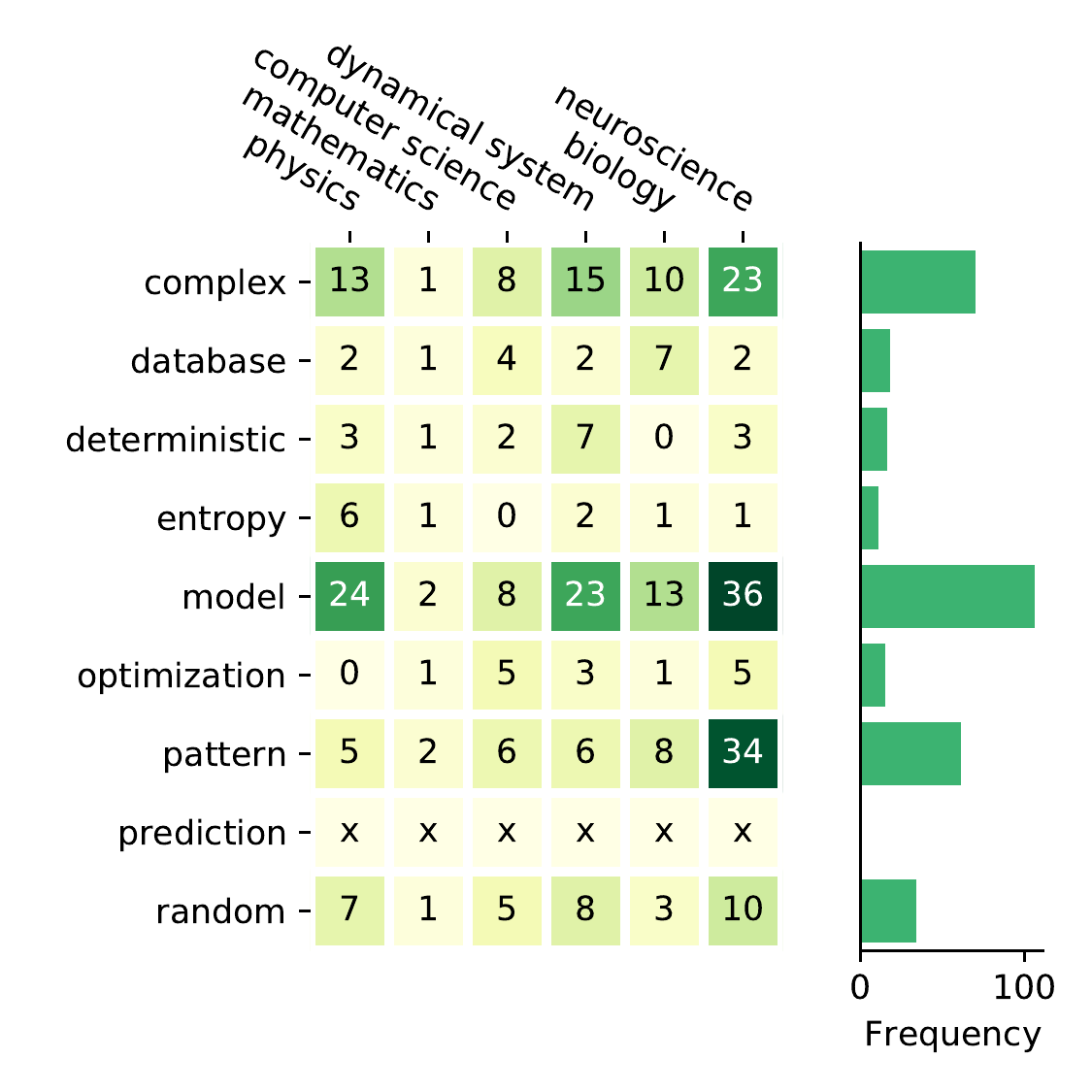}}
    \subfigure[\ random]{\includegraphics[width=0.325\textwidth]{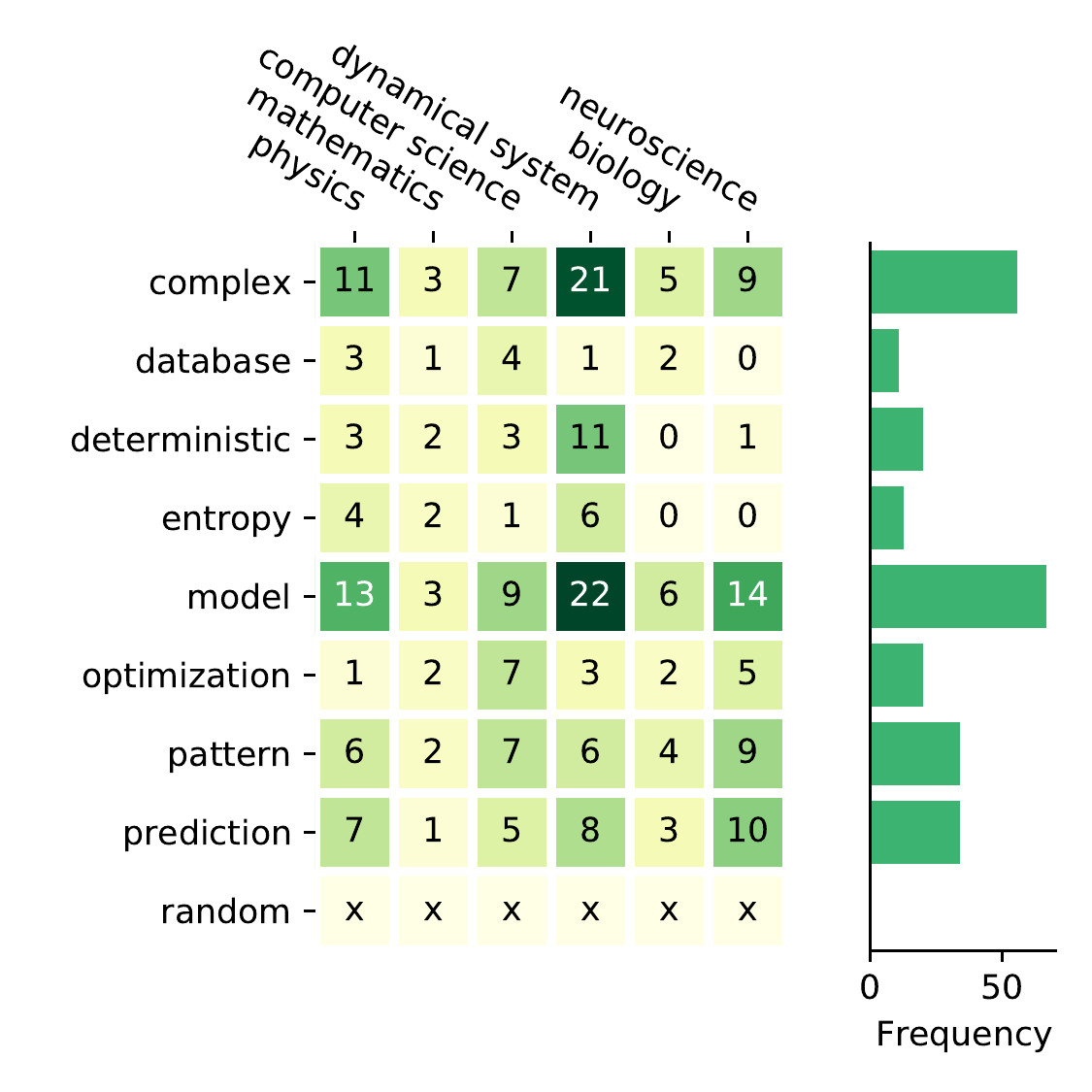}}
  \caption{Number of co-occurrence of pairs of words for each field. 
   The lateral bar plot represents the sum of the co-occurrences along the areas.}
  \label{fig:res_matrices}
\end{figure*}

As the next step of the analysis, we compare the co-occurrences within the documents for each of the analyzed areas.  Figure~\ref{fig:res_matrices} displays these frequencies for each word separately. As well as in the previous results, the word \emph{model} is found to co-occur more frequently. Many results presented here are in accordance with the discussion presented in Section~\ref{sec:words}. For example, \emph{complex} is strongly related to \emph{model} (see Figure~\ref{fig:res_matrices}(a)).  In contrast, \emph{complex} is not found to co-occur with \emph{entropy} for the analyzed areas. 

Interestingly, the word \emph{database} frequently co-occurs with \emph{complex} and \emph{model} in the areas of computer science and biology, as shown in Figure~\ref{fig:res_matrices}(b). For biology, these co-occurrences indicate a less straightforward relationship, highlighting the importance of databases for the development of biology.   When considering dynamical systems, as expected, the word \emph{deterministic} is found to co-occur with \emph{model}, as illustrated in Figure~\ref{fig:res_matrices}(c). The word \emph{deterministic} also co-occurs with \emph{complex}, which can be related to studies regarding complex systems modeling.

Furthermore, except for mathematics, \emph{model} is found to be related to complex (see Figure~\ref{fig:res_matrices}(e)). 
As shown in Figure~\ref{fig:res_matrices}(d), \emph{entropy} is not a frequent word in the considered areas. However, for the area of physics, this word is found to be related to \emph{model}. The word \emph{optimization} plays an important role in computer science and co-occur more frequently with \emph{complex}, \emph{model}, and \emph{pattern} (see Figure~\ref{fig:res_matrices}(f)). 

The area of neuroscience contains relatively more co-occurrences between \emph{pattern} and the words \emph{complex} and \emph{model}
(see Figure~\ref{fig:res_matrices}(g)). Similarly to this result, relationships between the term \emph{prediction} and the words \emph{model} and \emph{pattern} are also found in neuroscience, see Figure~\ref{fig:res_matrices}(h). As illustrated in Figure~\ref{fig:res_matrices}(i), the word random is frequently used in dynamical systems, and always co-occur with \emph{complex} and \emph{model}. This result may be related to the fact that, in the dynamical systems, potentially complex models are simulated by considering random variables. 

All in all, the obtained results support the fact that a same scientific word can be used quite distinctly in different areas, reflecting their current priorities and specific types of data, methods and modeling.

\subsection{Words colocalization network}
\label{sec:res_words_net}
\begin{figure*}[!ht]
  \centering
    \subfigure[\ Biology]{\includegraphics[width=0.325\textwidth]{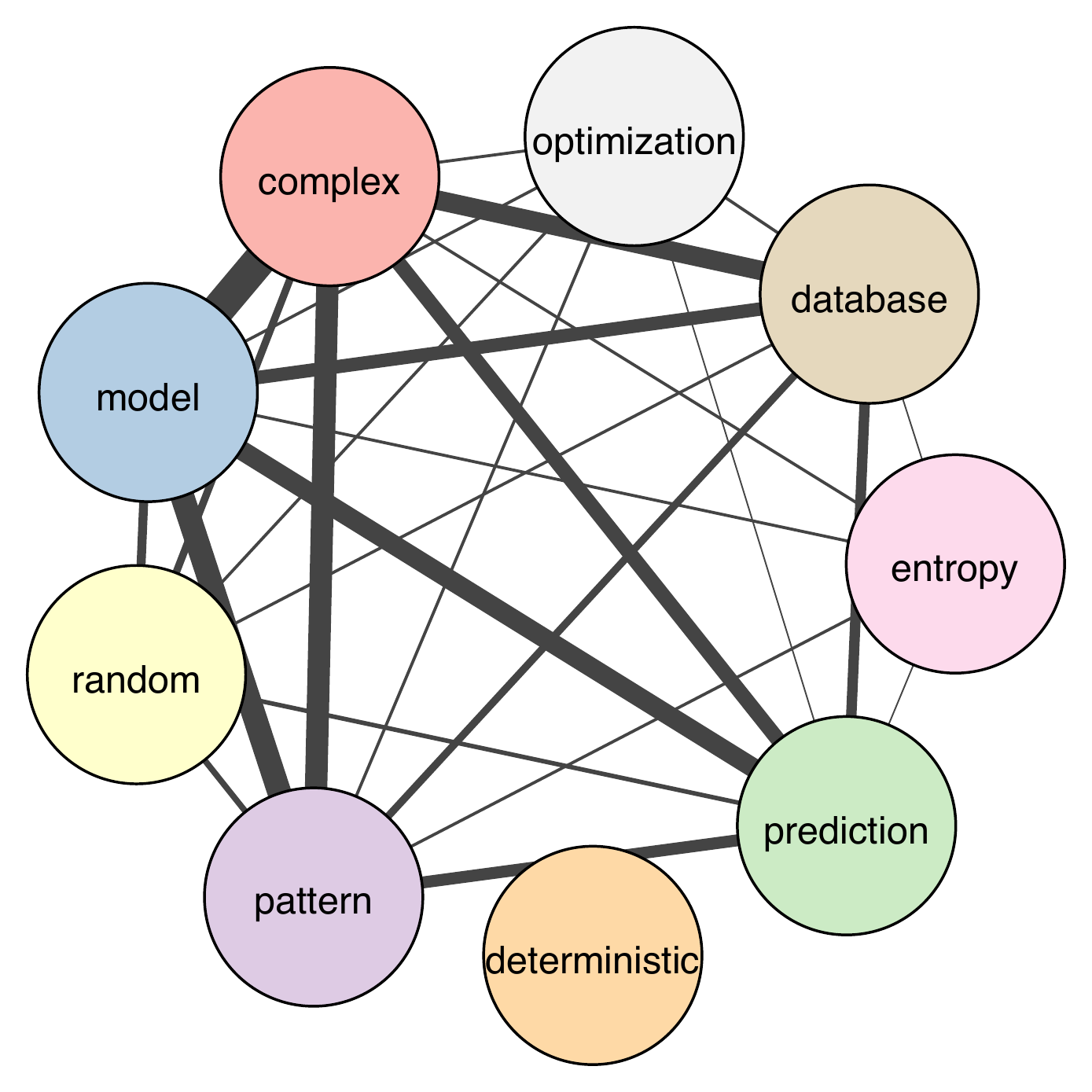}}
    \subfigure[\ Neuroscience]{\includegraphics[width=0.325\textwidth]{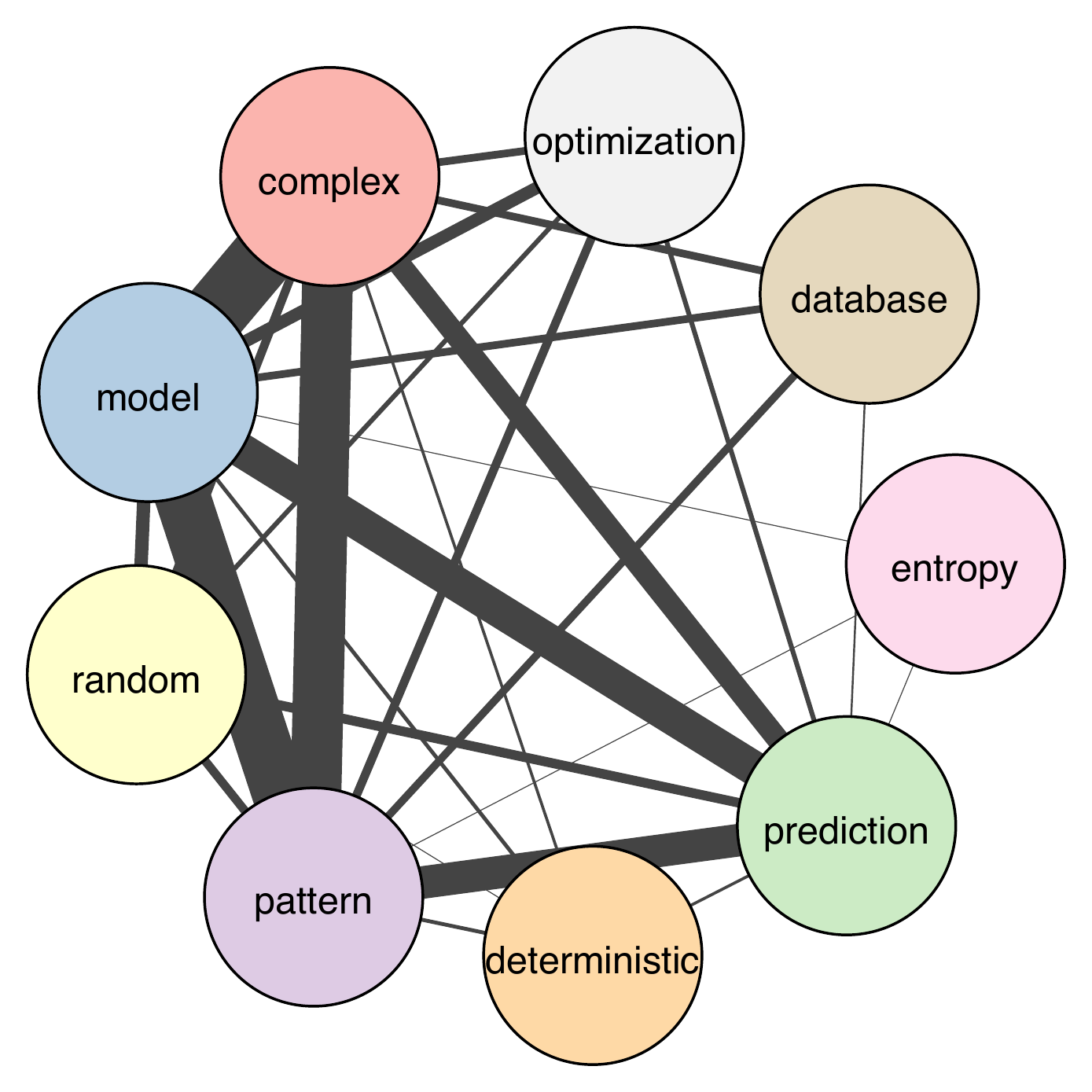}}

    \ 
    
    \subfigure[\ Physics]{\includegraphics[width=0.325\textwidth]{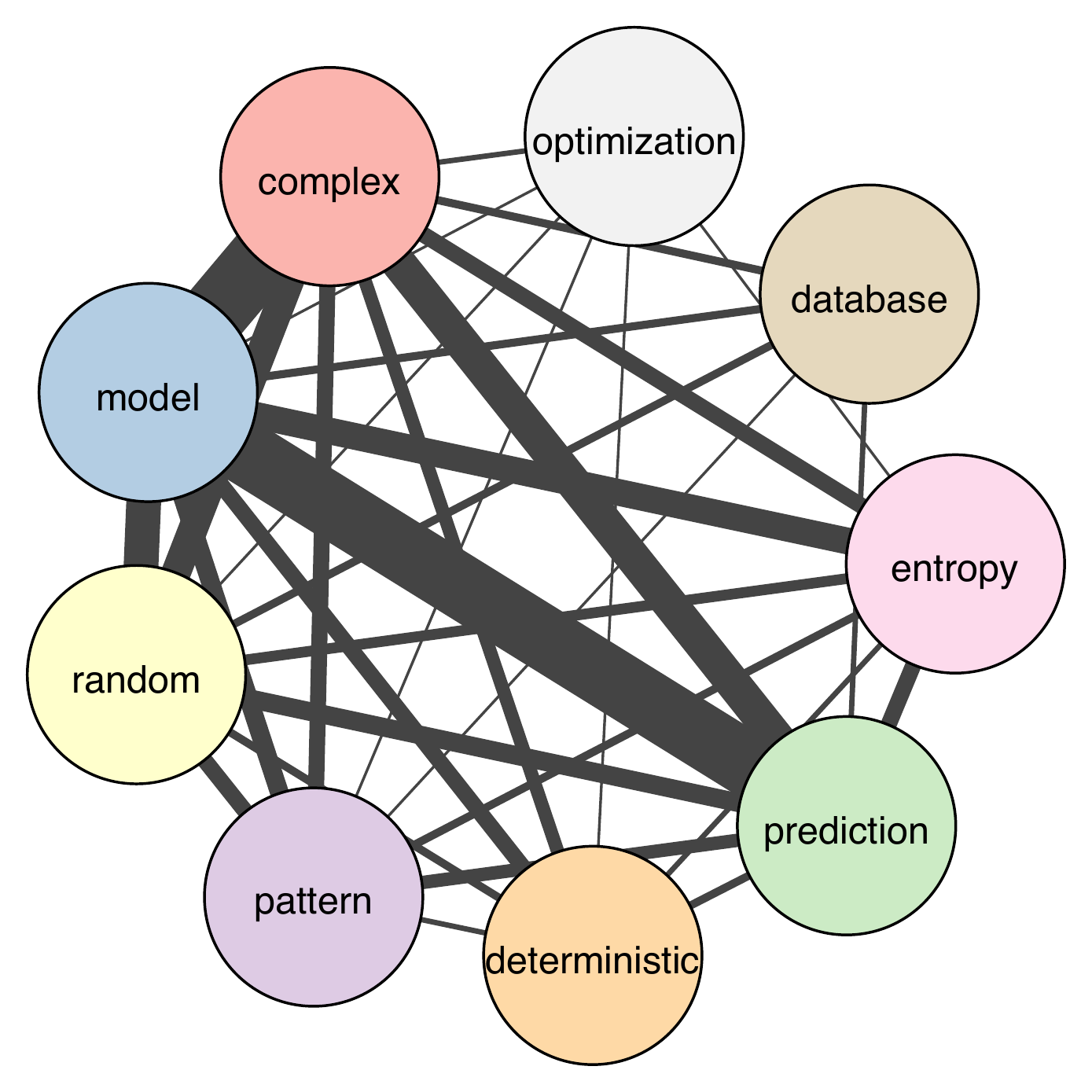}}
    \subfigure[\ Dynamical Systems]{\includegraphics[width=0.325\textwidth]{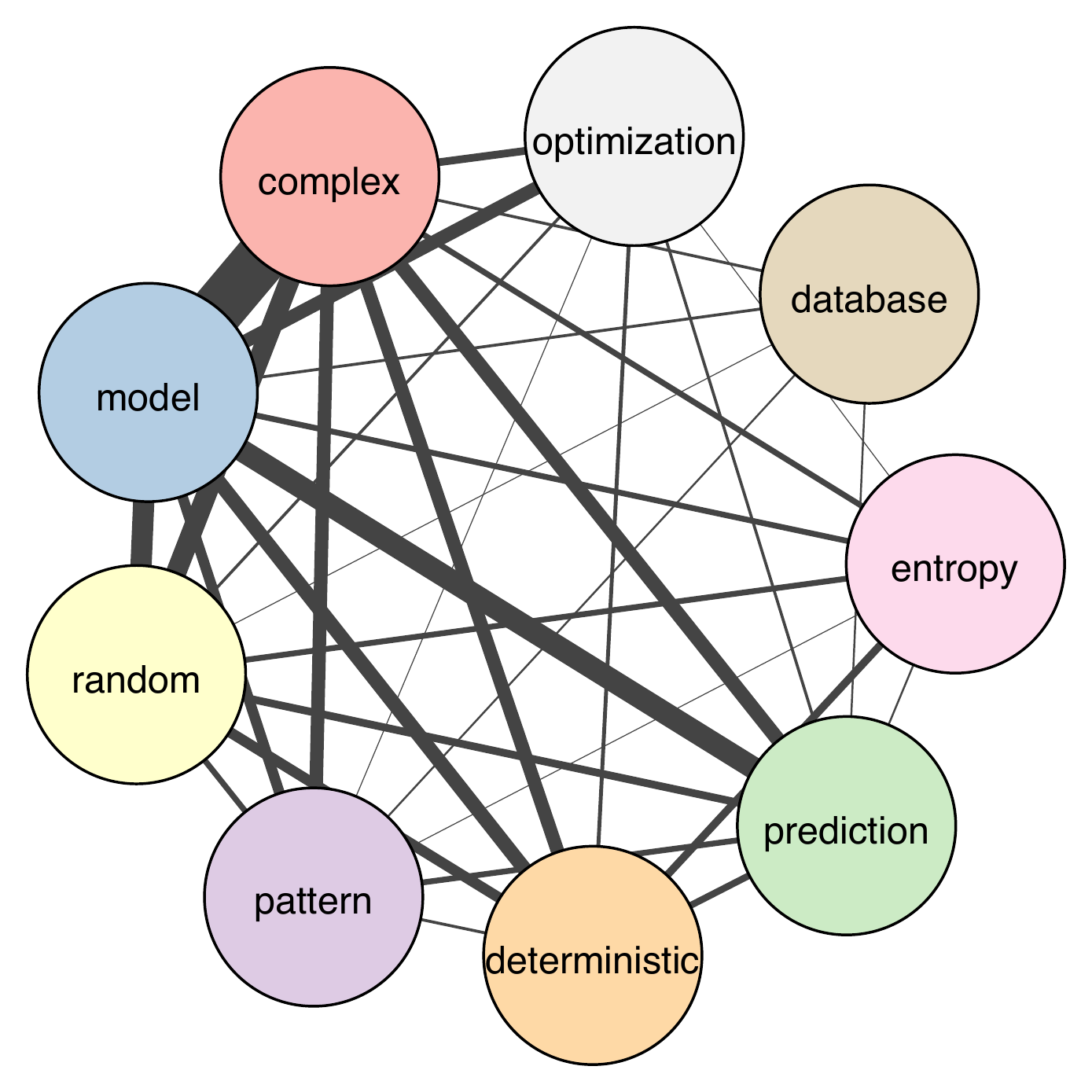}}

    \ 
    
    \subfigure[\ Mathematics]{\includegraphics[width=0.325\textwidth]{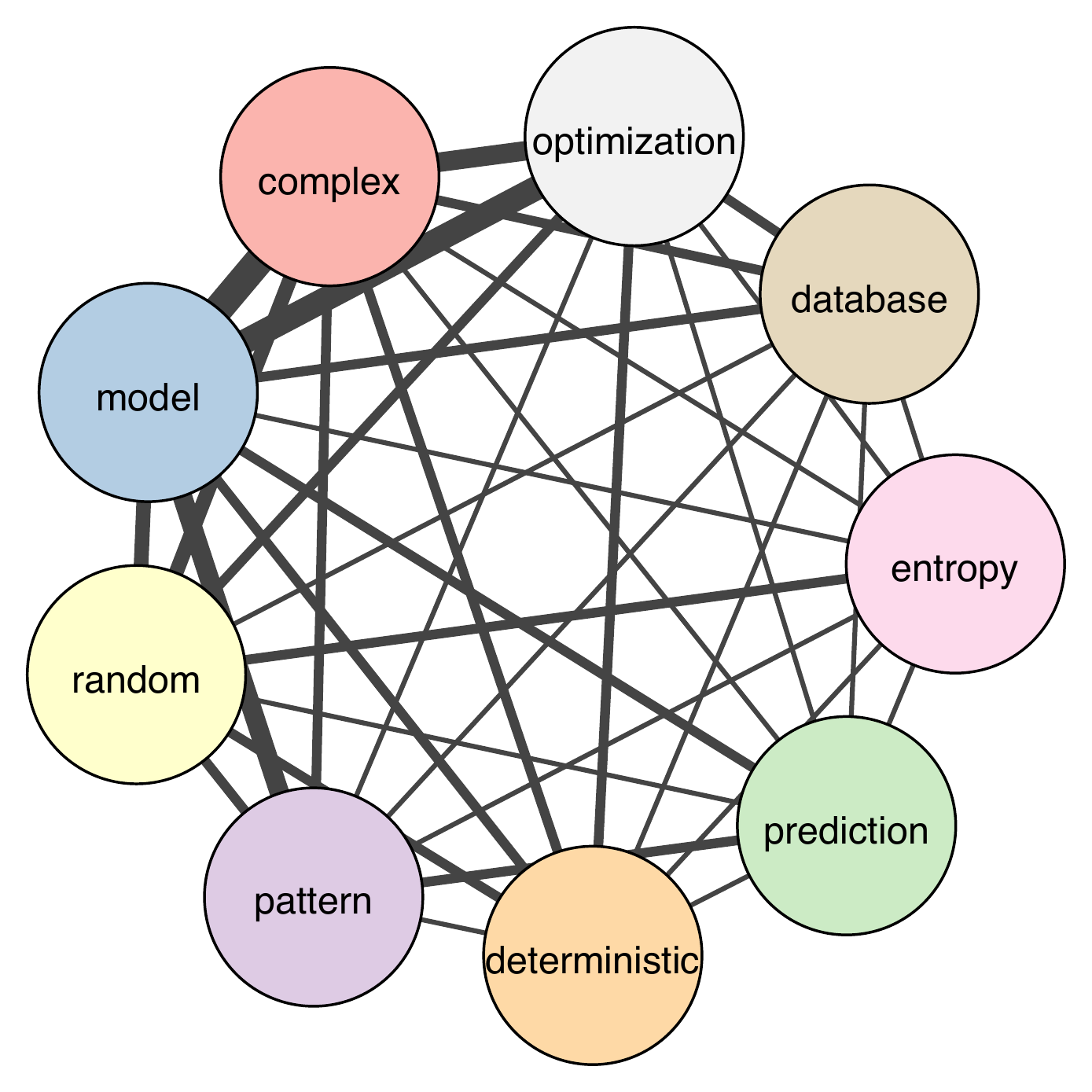}}
    \subfigure[\ Computer Science]{\includegraphics[width=0.325\textwidth]{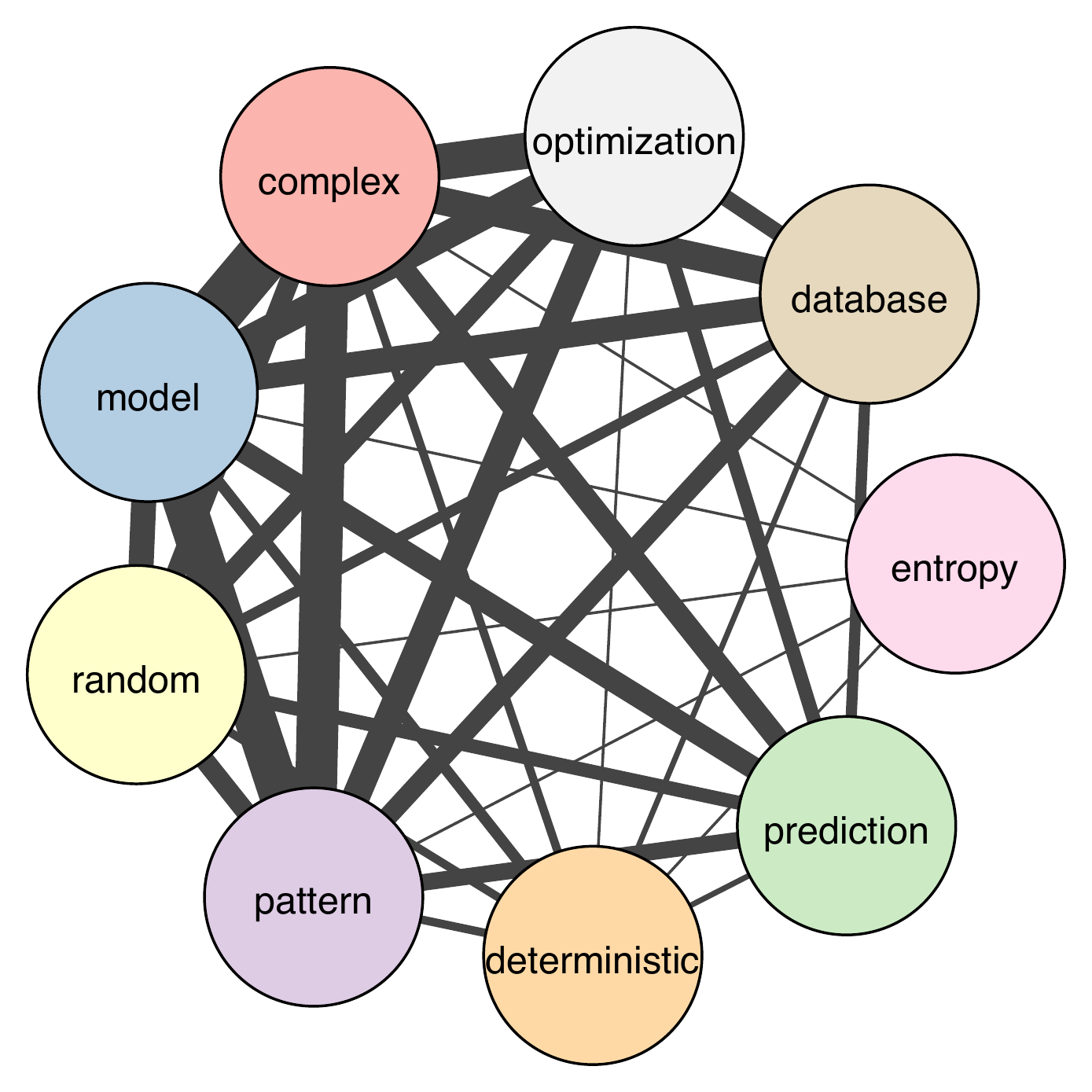}}
  \caption{Colocalization networks, in which the nodes represent the adopted words, and the edge widths the percentage of documents in which the pair of words co-occur. The edge connecting ``model'' and ``complex'' in item (c) corresponds to the maximum obtained co-occurrence, which represents 25\% of the articles. }
  \label{fig:res_words_relationship}
\end{figure*}

Further complementary analysis can be obtained in terms of the relationships between words obtained for each area. In this case, we employ the words colocalization network proposed in Section~\ref{sec:words_net}. The visualization of this network considering all fields is shown in Figure~\ref{fig:res_words_relationship}. As in the analysis developed in the previous section, the network visualizations display a wide range of connectivity patterns. This result indicates that, for all analyzed fields, the word contexts can differ. For instance, the weights of the connections between \emph{model} and \emph{prediction} present a pronounced variation. One strong relationship that was found to be preserved for all networks is between the words \emph{model} and \emph{complex}, indicating that complexity can be related to models regardless of the analyzed area. 

\subsection{Core relationships}
In addition to considering pairwise relationships between words, it is also interesting
to take into account the subset of words defined by the more intense relationships.
Henceforth, we understand the nodes interconnected by the 6 edges with largest weights
as corresponding to the core of each considered area.  For instance, in Figure\ref{fig:res_words_relationship}(b), which presents the relationships between the scientific words in the area of neuroscience, the words ``complex'', ``model'', ``random'', ``pattern'' and ``prediction'' are more strongly interconnected, giving rise to a respective \emph{core}. 

The analysis of these cores provide an interesting perspective from which to better understand
the different use and meaning of the considered words in distinct scientific areas, as well as
insights about the main current tendencies in each of those areas.  In the remainder of this 
section, we present a discussion about the cores obtained for each area taken into account,
which was developed taking into account information about the involved entries.

In the case of \emph{biology}, we have its core corresponding to the words ``complex'', ``model'', ``pattern'', ``prediction'', and ``database''.  These areas are strongly related to the
interface between \emph{biology} and \emph{computer science}, being directly associated to the multidisciplinary
areas of bioinformatics and systems biology.  These words can also be subdivided into two groups,
namely ``complex'', ``model'' and ``prediction'', which are closely related to modeling and the scientific method, and ``pattern'' and ``database'', which are characteristic of pattern analysis and
data science.  The union of these two subgroups are part of the scaffolding of bioinformatics
and systems biology.

The area of \emph{neuroscience} has a similar core to that found for \emph{biology}, involving the words ``complex'', ``model'', ``random'', ``pattern'' and ``prediction'', with the
word ``database'' being swapped to ``random''.  This seems to reflect the fact that randomness
is relatively more frequently associated to the other core words than the database.
Also, observe that the pattern of interconnection between these words is distinct from the previous case, with the words ``complex'', ``model'', and ``pattern'' being more strongly interrelated.  
To a great extent, this core also reflects the interface between neuroscience and computer science, being particularly related to the multidisciplinary areas of neuroinformatics and neurorobotics and
placing more emphasis on randomness as compared to database.  The identification of this
particular core indicates that these words are more typically used in associations related to the
above discussed interdisciplinary concepts.

The area of \emph{physics} has its core defied by the words ``complex'', ``model'', ``random'', ``prediction'', and ``entropy''.  This core is very similar to those identified for biology 
and neuroscience, with the difference that the word ``pattern'' as being replaced by ``entropy''. 
Again we have a close interrelationship between words directly related to modeling, namely ``complex'', ``model'', and ``prediction'', but now entropy assumes a more prominent role, reflecting the importance of this concept in the area of physics.

The core identified for the area of \emph{dynamical system} includes the words  ``complex'', ``model'', ``random'', ``deterministic'', and ``prediction''.  The words ``complex'', ``model'', and ``prediction'' again indicate a 
strong emphasis on modeling and the scientific method, but now we have a new substantial
relationship also with the word ``deterministic'', which indicates that the other words are
often being employed while discussing the key issue of a system being (or not) deterministic, which
is reasonable given the importance of this concept in the area of \emph{dynamic systems}.

The area of \emph{mathematics} resulted with respective core incorporating the words
``complex'', ``model'', ``random'',  ``pattern'', and ``optimization''.  Three words present particularly strong interrelationship, namely ``complex'', ``model'', and ``optimization''.  This indicates emphasis on the concept of optimization, which is indeed a recurrent subject in 
mathematics.  The word ``pattern'' tends to be associated with the word model, indicating that
its main usage could be related to the complexity of patterns in abstract and physical worlds.
    
The core obtained for the area of \emph{computer Science} includes the words.
 ``complex'', ``random'', ``pattern'', ``database'', and ``optimization''.  In particular, we
 observe the absence of the word ``model'', which appears in all the other areas considered in
 the present work.  Again, we observe that the world complex is particularly associated to
 ``pattern'', which could be related to areas such as computational complexity and pattern
 generation or recognition.  ``Optimization'' also presents a strong relationship with the
 word ``complex'', suggesting associations with the area of computational optimization.

\subsection{Summarized network}
\begin{figure}[!ht]
  \centering
    \includegraphics[width=0.325\textwidth]{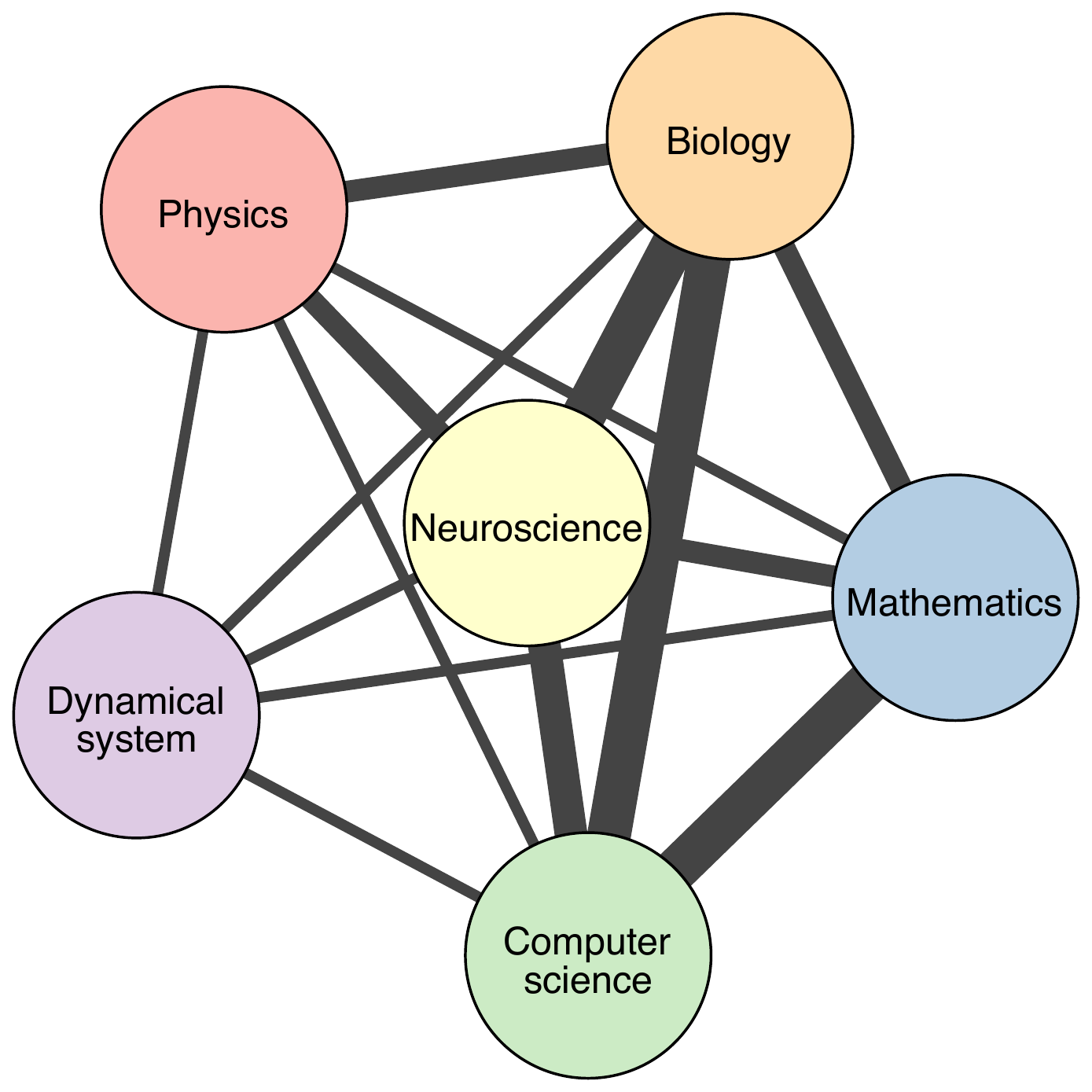}
\caption{Summarized network, representing the relationships between all the networks in Figure~\ref{fig:res_words_relationship}.  Observe that each node in this network corresponds to a whole respective networks in the latter figure, while the edges
were obtained considering the similarity of connections as described in Section~\ref{sec:summary_net}.}
  \label{fig:net_summary_areas}
\end{figure}

In order to better understand the relationship among the networks presented in Figure~\ref{fig:res_words_relationship}, we created the summarized network (see Figure~\ref{fig:net_summary_areas}). As can be seen, the relationships between the analyzed terms vary. The edge between some pairs of fields are not shown because that relationship is too small, which reinforces the idea that the words tend to be employed in different manners for each analyzed field.  

\section{\label{sec:conc} Conclusions}

In writing and speaking, words are used to communicate insights and pieces of information.  However, a same word can have different meanings depending on its context, such as the area in which it is employed.  In order to better understand this interesting phenomenon, we study some particularly some relevant scientific words (``prediction'', ``model'', ``optimization'', ``complex'', ``entropy'', ``random'', ``deterministic'', ``pattern'', and ``database'') respectively to some major areas of study (``biology'', ``neuroscience'', ``physics'', ``dynamical systems'', ``mathematics'', and ``computer science''). These areas were analyzed by considering pages obtained from Wikipedia.

Aiming at understanding the relationship between the selected scientific words, we considered a network where the nodes represent words and the edges are weighted according to the colocalization of the words in the entire entries, i.e.~when two words appear in a same Wikipedia page.

Many interesting results were obtained. In general, the relationships between fields were shown.  For example, the word ``prediction'' was found to be related to ``model'' and ``complex''. However, the relationship with ``random'' was not observed.  Relationships between ``complex'' and ``database'' were also expected for most fields. However, it was found only for the areas of \emph{computer science} and \emph{biology}. As expected, the word ``model'' turned out to be related to ``database'' in \emph{biology} and \emph{neuroscience}, but these relationships are not too strong in comparison with others, mainly for \emph{neuroscience}. The word ``optimization'' was found to play an important role in the area of \emph{computer science}. This word was also observed to be strong in the area of \emph{mathematics} but, in this case, the contexts are different. The relationship between ``entropy'' and ``complex'' and ``random'' was found for the field of \emph{physics}. Since the word ``entropy'' is intrinsically related to information theory, we expected that this word could play an important role in \emph{computer science}, which was not the case. For all considered fields, the word ``deterministic'' was not observe to be strongly related to other words but, as supposed, though it resulted being moderately related to ``random''. Many relationships could also be expected to be obtained between the word ``pattern'' and the words ``model'', ``random'', ``complex'', and ``database''. In the case of the latter, its relationship with the word ``pattern'' was verified only for the area of \emph{computer science}.  

By defining a core for each of the colocalization networks, it was possible to discuss relationships between the considered scientific areas.
Surprisingly, the areas of \emph{biology} and \emph{computer science} were found to be similar when considering the relationships of the words ``database'' and ``prediction''. This result illustrates the proximity between these fields, which relates to the field of bioinformatics.  Furthermore, the relationships indicate that these words tend to have similar meanings in both areas, which allows the researchers to communicate with particular effectiveness among themselves. 

Taking into account that the context is essential to the solution of tasks related to natural language processing (e.g., text classification, authorship attribution, and sentiment analysis), our study paves the way for future studies. As future works, other scientific words and fields can be studied. Furthermore, our analysis can be considered in order to incorporate more relevant information, such as in embedding techniques. 

\section*{Acknowledgments}
Henrique F. de Arruda acknowledges FAPESP for sponsorship (grant no. 2018/10489-0). Luciano da F. Costa thanks CNPq (grant no. 307085/2018-0) for sponsorship. This work has been supported also by FAPESP grant no. 2015/22308-2.

\bibliography{ref}

\begin{thebibliography}{39}%
\makeatletter
\providecommand \@ifxundefined [1]{%
 \@ifx{#1\undefined}
}%
\providecommand \@ifnum [1]{%
 \ifnum #1\expandafter \@firstoftwo
 \else \expandafter \@secondoftwo
 \fi
}%
\providecommand \@ifx [1]{%
 \ifx #1\expandafter \@firstoftwo
 \else \expandafter \@secondoftwo
 \fi
}%
\providecommand \natexlab [1]{#1}%
\providecommand \enquote  [1]{``#1''}%
\providecommand \bibnamefont  [1]{#1}%
\providecommand \bibfnamefont [1]{#1}%
\providecommand \citenamefont [1]{#1}%
\providecommand \href@noop [0]{\@secondoftwo}%
\providecommand \href [0]{\begingroup \@sanitize@url \@href}%
\providecommand \@href[1]{\@@startlink{#1}\@@href}%
\providecommand \@@href[1]{\endgroup#1\@@endlink}%
\providecommand \@sanitize@url [0]{\catcode `\\12\catcode `\$12\catcode
  `\&12\catcode `\#12\catcode `\^12\catcode `\_12\catcode `\%12\relax}%
\providecommand \@@startlink[1]{}%
\providecommand \@@endlink[0]{}%
\providecommand \url  [0]{\begingroup\@sanitize@url \@url }%
\providecommand \@url [1]{\endgroup\@href {#1}{\urlprefix }}%
\providecommand \urlprefix  [0]{URL }%
\providecommand \Eprint [0]{\href }%
\providecommand \doibase [0]{http://dx.doi.org/}%
\providecommand \selectlanguage [0]{\@gobble}%
\providecommand \bibinfo  [0]{\@secondoftwo}%
\providecommand \bibfield  [0]{\@secondoftwo}%
\providecommand \translation [1]{[#1]}%
\providecommand \BibitemOpen [0]{}%
\providecommand \bibitemStop [0]{}%
\providecommand \bibitemNoStop [0]{.\EOS\space}%
\providecommand \EOS [0]{\spacefactor3000\relax}%
\providecommand \BibitemShut  [1]{\csname bibitem#1\endcsname}%
\let\auto@bib@innerbib\@empty
\bibitem [{\citenamefont {Rosvall}\ and\ \citenamefont
  {Bergstrom}(2011)}]{rosvall2011multilevel}%
  \BibitemOpen
  \bibfield  {author} {\bibinfo {author} {\bibfnamefont {M.}~\bibnamefont
  {Rosvall}}\ and\ \bibinfo {author} {\bibfnamefont {C.~T.}\ \bibnamefont
  {Bergstrom}},\ }\href@noop {} {\bibfield  {journal} {\bibinfo  {journal}
  {PloS one}\ }\textbf {\bibinfo {volume} {6}},\ \bibinfo {pages} {e18209}
  (\bibinfo {year} {2011})}\BibitemShut {NoStop}%
\bibitem [{\citenamefont {Morillo}\ \emph {et~al.}(2003)\citenamefont
  {Morillo}, \citenamefont {Bordons},\ and\ \citenamefont
  {G{\'o}mez}}]{morillo2003interdisciplinarity}%
  \BibitemOpen
  \bibfield  {author} {\bibinfo {author} {\bibfnamefont {F.}~\bibnamefont
  {Morillo}}, \bibinfo {author} {\bibfnamefont {M.}~\bibnamefont {Bordons}}, \
  and\ \bibinfo {author} {\bibfnamefont {I.}~\bibnamefont {G{\'o}mez}},\
  }\href@noop {} {\bibfield  {journal} {\bibinfo  {journal} {Journal of the
  American Society for Information Science and technology}\ }\textbf {\bibinfo
  {volume} {54}},\ \bibinfo {pages} {1237} (\bibinfo {year}
  {2003})}\BibitemShut {NoStop}%
\bibitem [{\citenamefont {Small}(2006)}]{small2006tracking}%
  \BibitemOpen
  \bibfield  {author} {\bibinfo {author} {\bibfnamefont {H.}~\bibnamefont
  {Small}},\ }\href@noop {} {\bibfield  {journal} {\bibinfo  {journal}
  {Scientometrics}\ }\textbf {\bibinfo {volume} {68}},\ \bibinfo {pages} {595}
  (\bibinfo {year} {2006})}\BibitemShut {NoStop}%
\bibitem [{\citenamefont {Whittaker}(1989)}]{whittaker1989creativity}%
  \BibitemOpen
  \bibfield  {author} {\bibinfo {author} {\bibfnamefont {J.}~\bibnamefont
  {Whittaker}},\ }\href@noop {} {\bibfield  {journal} {\bibinfo  {journal}
  {Social Studies of Science}\ }\textbf {\bibinfo {volume} {19}},\ \bibinfo
  {pages} {473} (\bibinfo {year} {1989})}\BibitemShut {NoStop}%
\bibitem [{\citenamefont {Sicilia-Garcia}\ \emph {et~al.}(2003)\citenamefont
  {Sicilia-Garcia}, \citenamefont {Ming}, \citenamefont {Smith} \emph
  {et~al.}}]{sicilia2003extension}%
  \BibitemOpen
  \bibfield  {author} {\bibinfo {author} {\bibfnamefont {E.~I.}\ \bibnamefont
  {Sicilia-Garcia}}, \bibinfo {author} {\bibfnamefont {J.}~\bibnamefont
  {Ming}}, \bibinfo {author} {\bibfnamefont {F.~J.}\ \bibnamefont {Smith}},
  \emph {et~al.},\ }in\ \href@noop {} {\emph {\bibinfo {booktitle}
  {International Journal of Computational Linguistics \& Chinese Language
  Processing, Volume 8, Number 1, February 2003: Special Issue on Word
  Formation and Chinese Language Processing}}}\ (\bibinfo {year} {2003})\ pp.\
  \bibinfo {pages} {77--102}\BibitemShut {NoStop}%
\bibitem [{\citenamefont {Cavnar}\ \emph {et~al.}(1994)\citenamefont {Cavnar},
  \citenamefont {Trenkle} \emph {et~al.}}]{cavnar1994n}%
  \BibitemOpen
  \bibfield  {author} {\bibinfo {author} {\bibfnamefont {W.~B.}\ \bibnamefont
  {Cavnar}}, \bibinfo {author} {\bibfnamefont {J.~M.}\ \bibnamefont {Trenkle}},
   \emph {et~al.},\ }in\ \href@noop {} {\emph {\bibinfo {booktitle}
  {Proceedings of SDAIR-94, 3rd annual symposium on document analysis and
  information retrieval}}},\ Vol.\ \bibinfo {volume} {161175}\ (\bibinfo
  {organization} {Citeseer},\ \bibinfo {year} {1994})\BibitemShut {NoStop}%
\bibitem [{\citenamefont {Tripathy}\ \emph {et~al.}(2016)\citenamefont
  {Tripathy}, \citenamefont {Agrawal},\ and\ \citenamefont
  {Rath}}]{tripathy2016classification}%
  \BibitemOpen
  \bibfield  {author} {\bibinfo {author} {\bibfnamefont {A.}~\bibnamefont
  {Tripathy}}, \bibinfo {author} {\bibfnamefont {A.}~\bibnamefont {Agrawal}}, \
  and\ \bibinfo {author} {\bibfnamefont {S.~K.}\ \bibnamefont {Rath}},\
  }\href@noop {} {\bibfield  {journal} {\bibinfo  {journal} {Expert Systems
  with Applications}\ }\textbf {\bibinfo {volume} {57}},\ \bibinfo {pages}
  {117} (\bibinfo {year} {2016})}\BibitemShut {NoStop}%
\bibitem [{\citenamefont {Khreisat}(2006)}]{khreisat2006arabic}%
  \BibitemOpen
  \bibfield  {author} {\bibinfo {author} {\bibfnamefont {L.}~\bibnamefont
  {Khreisat}},\ }\href@noop {} {\bibfield  {journal} {\bibinfo  {journal}
  {DMIN}\ }\textbf {\bibinfo {volume} {2006}},\ \bibinfo {pages} {78} (\bibinfo
  {year} {2006})}\BibitemShut {NoStop}%
\bibitem [{\citenamefont {Peng}\ and\ \citenamefont
  {Schuurmans}(2003)}]{peng2003combining}%
  \BibitemOpen
  \bibfield  {author} {\bibinfo {author} {\bibfnamefont {F.}~\bibnamefont
  {Peng}}\ and\ \bibinfo {author} {\bibfnamefont {D.}~\bibnamefont
  {Schuurmans}},\ }in\ \href@noop {} {\emph {\bibinfo {booktitle} {European
  Conference on Information Retrieval}}}\ (\bibinfo {organization} {Springer},\
  \bibinfo {year} {2003})\ pp.\ \bibinfo {pages} {335--350}\BibitemShut
  {NoStop}%
\bibitem [{\citenamefont {Khreisat}(2009)}]{khreisat2009machine}%
  \BibitemOpen
  \bibfield  {author} {\bibinfo {author} {\bibfnamefont {L.}~\bibnamefont
  {Khreisat}},\ }\href@noop {} {\bibfield  {journal} {\bibinfo  {journal}
  {Journal of Informetrics}\ }\textbf {\bibinfo {volume} {3}},\ \bibinfo
  {pages} {72} (\bibinfo {year} {2009})}\BibitemShut {NoStop}%
\bibitem [{\citenamefont {Al-Shalabi}\ and\ \citenamefont
  {Obeidat}(2008)}]{al2008improving}%
  \BibitemOpen
  \bibfield  {author} {\bibinfo {author} {\bibfnamefont {R.}~\bibnamefont
  {Al-Shalabi}}\ and\ \bibinfo {author} {\bibfnamefont {R.}~\bibnamefont
  {Obeidat}},\ }in\ \href@noop {} {\emph {\bibinfo {booktitle} {Proceedings of
  the Sixth International Conference on Informatics and Systems, Cairo,
  Egypt}}}\ (\bibinfo {organization} {Citeseer},\ \bibinfo {year} {2008})\ pp.\
  \bibinfo {pages} {108--112}\BibitemShut {NoStop}%
\bibitem [{\citenamefont {Zampieri}(2013)}]{zampieri2013using}%
  \BibitemOpen
  \bibfield  {author} {\bibinfo {author} {\bibfnamefont {M.}~\bibnamefont
  {Zampieri}},\ }in\ \href@noop {} {\emph {\bibinfo {booktitle} {2013 IEEE 14th
  international symposium on computational intelligence and informatics
  (CINTI)}}}\ (\bibinfo {organization} {IEEE},\ \bibinfo {year} {2013})\ pp.\
  \bibinfo {pages} {37--41}\BibitemShut {NoStop}%
\bibitem [{\citenamefont {Silva}\ \emph {et~al.}(2016)\citenamefont {Silva},
  \citenamefont {Amancio}, \citenamefont {Bardosova}, \citenamefont
  {da~F.~Costa},\ and\ \citenamefont {Oliveira~Jr}}]{silva2016using}%
  \BibitemOpen
  \bibfield  {author} {\bibinfo {author} {\bibfnamefont {F.~N.}\ \bibnamefont
  {Silva}}, \bibinfo {author} {\bibfnamefont {D.~R.}\ \bibnamefont {Amancio}},
  \bibinfo {author} {\bibfnamefont {M.}~\bibnamefont {Bardosova}}, \bibinfo
  {author} {\bibfnamefont {L.}~\bibnamefont {da~F.~Costa}}, \ and\ \bibinfo
  {author} {\bibfnamefont {O.~N.}\ \bibnamefont {Oliveira~Jr}},\ }\href@noop {}
  {\bibfield  {journal} {\bibinfo  {journal} {Journal of Informetrics}\
  }\textbf {\bibinfo {volume} {10}},\ \bibinfo {pages} {487} (\bibinfo {year}
  {2016})}\BibitemShut {NoStop}%
\bibitem [{\citenamefont {Ceribeli}\ \emph {et~al.}(2021)\citenamefont
  {Ceribeli}, \citenamefont {de~Arruda},\ and\ \citenamefont
  {da~F.~Costa}}]{ceribeli2021coupled}%
  \BibitemOpen
  \bibfield  {author} {\bibinfo {author} {\bibfnamefont {C.}~\bibnamefont
  {Ceribeli}}, \bibinfo {author} {\bibfnamefont {H.~F.}\ \bibnamefont
  {de~Arruda}}, \ and\ \bibinfo {author} {\bibfnamefont {L.}~\bibnamefont
  {da~F.~Costa}},\ }\href@noop {} {\bibfield  {journal} {\bibinfo  {journal}
  {Scientometrics}\ }\textbf {\bibinfo {volume} {126}},\ \bibinfo {pages}
  {3841–3851} (\bibinfo {year} {2021})}\BibitemShut {NoStop}%
\bibitem [{\citenamefont {Benatti}\ \emph {et~al.}(2021)\citenamefont
  {Benatti}, \citenamefont {de~Arruda}, \citenamefont {Silva},\ and\
  \citenamefont {da~F.~Costa}}]{benatti2021enriching}%
  \BibitemOpen
  \bibfield  {author} {\bibinfo {author} {\bibfnamefont {A.}~\bibnamefont
  {Benatti}}, \bibinfo {author} {\bibfnamefont {H.~F.}\ \bibnamefont
  {de~Arruda}}, \bibinfo {author} {\bibfnamefont {F.~N.}\ \bibnamefont
  {Silva}}, \ and\ \bibinfo {author} {\bibfnamefont {L.}~\bibnamefont
  {da~F.~Costa}},\ }\href@noop {} {\bibfield  {journal} {\bibinfo  {journal}
  {Physica A: Statistical Mechanics and its Applications}\ }\textbf {\bibinfo
  {volume} {573}},\ \bibinfo {pages} {125901} (\bibinfo {year}
  {2021})}\BibitemShut {NoStop}%
\bibitem [{\citenamefont {Brede}\ and\ \citenamefont
  {Newth}(2008)}]{brede2008patterns}%
  \BibitemOpen
  \bibfield  {author} {\bibinfo {author} {\bibfnamefont {M.}~\bibnamefont
  {Brede}}\ and\ \bibinfo {author} {\bibfnamefont {D.}~\bibnamefont {Newth}},\
  }\href@noop {} {\bibfield  {journal} {\bibinfo  {journal} {Complexity
  International}\ }\textbf {\bibinfo {volume} {12}} (\bibinfo {year}
  {2008})}\BibitemShut {NoStop}%
\bibitem [{\citenamefont {Amancio}\ \emph {et~al.}(2012)\citenamefont
  {Amancio}, \citenamefont {Oliveira~Jr},\ and\ \citenamefont
  {da~F.~Costa}}]{amancio2012identification}%
  \BibitemOpen
  \bibfield  {author} {\bibinfo {author} {\bibfnamefont {D.~R.}\ \bibnamefont
  {Amancio}}, \bibinfo {author} {\bibfnamefont {O.~N.}\ \bibnamefont
  {Oliveira~Jr}}, \ and\ \bibinfo {author} {\bibfnamefont {L.}~\bibnamefont
  {da~F.~Costa}},\ }\href@noop {} {\bibfield  {journal} {\bibinfo  {journal}
  {New Journal of Physics}\ }\textbf {\bibinfo {volume} {14}},\ \bibinfo
  {pages} {043029} (\bibinfo {year} {2012})}\BibitemShut {NoStop}%
\bibitem [{\citenamefont {de~Arruda}\ \emph {et~al.}(2016)\citenamefont
  {de~Arruda}, \citenamefont {da~F.~Costa},\ and\ \citenamefont
  {Amancio}}]{de2016using}%
  \BibitemOpen
  \bibfield  {author} {\bibinfo {author} {\bibfnamefont {H.~F.}\ \bibnamefont
  {de~Arruda}}, \bibinfo {author} {\bibfnamefont {L.}~\bibnamefont
  {da~F.~Costa}}, \ and\ \bibinfo {author} {\bibfnamefont {D.~R.}\ \bibnamefont
  {Amancio}},\ }\href@noop {} {\bibfield  {journal} {\bibinfo  {journal} {EPL
  (Europhysics Letters)}\ }\textbf {\bibinfo {volume} {113}},\ \bibinfo {pages}
  {28007} (\bibinfo {year} {2016})}\BibitemShut {NoStop}%
\bibitem [{\citenamefont {Marinho}\ \emph {et~al.}(2016)\citenamefont
  {Marinho}, \citenamefont {Hirst},\ and\ \citenamefont
  {Amancio}}]{marinho2016authorship}%
  \BibitemOpen
  \bibfield  {author} {\bibinfo {author} {\bibfnamefont {V.~Q.}\ \bibnamefont
  {Marinho}}, \bibinfo {author} {\bibfnamefont {G.}~\bibnamefont {Hirst}}, \
  and\ \bibinfo {author} {\bibfnamefont {D.~R.}\ \bibnamefont {Amancio}},\ }in\
  \href@noop {} {\emph {\bibinfo {booktitle} {2016 5th Brazilian Conference on
  Intelligent Systems (BRACIS)}}}\ (\bibinfo {organization} {IEEE},\ \bibinfo
  {year} {2016})\ pp.\ \bibinfo {pages} {355--360}\BibitemShut {NoStop}%
\bibitem [{\citenamefont {Hu}\ \emph {et~al.}(2016)\citenamefont {Hu},
  \citenamefont {Zhang},\ and\ \citenamefont {Zheng}}]{hu2016different}%
  \BibitemOpen
  \bibfield  {author} {\bibinfo {author} {\bibfnamefont {W.}~\bibnamefont
  {Hu}}, \bibinfo {author} {\bibfnamefont {J.}~\bibnamefont {Zhang}}, \ and\
  \bibinfo {author} {\bibfnamefont {N.}~\bibnamefont {Zheng}},\ }in\ \href@noop
  {} {\emph {\bibinfo {booktitle} {Proceedings of COLING 2016, the 26th
  International Conference on Computational Linguistics: Technical Papers}}}\
  (\bibinfo {year} {2016})\ pp.\ \bibinfo {pages} {762--771}\BibitemShut
  {NoStop}%
\bibitem [{\citenamefont {Pennington}\ \emph {et~al.}(2014)\citenamefont
  {Pennington}, \citenamefont {Socher},\ and\ \citenamefont
  {Manning}}]{pennington2014glove}%
  \BibitemOpen
  \bibfield  {author} {\bibinfo {author} {\bibfnamefont {J.}~\bibnamefont
  {Pennington}}, \bibinfo {author} {\bibfnamefont {R.}~\bibnamefont {Socher}},
  \ and\ \bibinfo {author} {\bibfnamefont {C.~D.}\ \bibnamefont {Manning}},\
  }in\ \href@noop {} {\emph {\bibinfo {booktitle} {Proceedings of the 2014
  conference on empirical methods in natural language processing (EMNLP)}}}\
  (\bibinfo {year} {2014})\ pp.\ \bibinfo {pages} {1532--1543}\BibitemShut
  {NoStop}%
\bibitem [{\citenamefont {Mikolov}\ \emph {et~al.}(2013)\citenamefont
  {Mikolov}, \citenamefont {Chen}, \citenamefont {Corrado},\ and\ \citenamefont
  {Dean}}]{mikolov2013efficient}%
  \BibitemOpen
  \bibfield  {author} {\bibinfo {author} {\bibfnamefont {T.}~\bibnamefont
  {Mikolov}}, \bibinfo {author} {\bibfnamefont {K.}~\bibnamefont {Chen}},
  \bibinfo {author} {\bibfnamefont {G.}~\bibnamefont {Corrado}}, \ and\
  \bibinfo {author} {\bibfnamefont {J.}~\bibnamefont {Dean}},\ }\href@noop {}
  {\bibfield  {journal} {\bibinfo  {journal} {arXiv preprint arXiv:1301.3781}\
  } (\bibinfo {year} {2013})}\BibitemShut {NoStop}%
\bibitem [{\citenamefont {da~F.~Costa}(2020)}]{costa2020learning}%
  \BibitemOpen
  \bibfield  {author} {\bibinfo {author} {\bibfnamefont {L.}~\bibnamefont
  {da~F.~Costa}},\ }\href {\doibase 10.13140/RG.2.2.29803.16160/2} {\
  (\bibinfo {year} {2020}),\ 10.13140/RG.2.2.29803.16160/2}\BibitemShut
  {NoStop}%
\bibitem [{\citenamefont {Brown}\ and\ \citenamefont
  {Br{\"u}ne}(2012)}]{brown2012role}%
  \BibitemOpen
  \bibfield  {author} {\bibinfo {author} {\bibfnamefont {E.~C.}\ \bibnamefont
  {Brown}}\ and\ \bibinfo {author} {\bibfnamefont {M.}~\bibnamefont
  {Br{\"u}ne}},\ }\href@noop {} {\bibfield  {journal} {\bibinfo  {journal}
  {Frontiers in human neuroscience}\ }\textbf {\bibinfo {volume} {6}},\
  \bibinfo {pages} {147} (\bibinfo {year} {2012})}\BibitemShut {NoStop}%
\bibitem [{\citenamefont {Stepanchikova}\ \emph {et~al.}(2003)\citenamefont
  {Stepanchikova}, \citenamefont {Lagunin}, \citenamefont {Filimonov},\ and\
  \citenamefont {Poroikov}}]{stepanchikova2003prediction}%
  \BibitemOpen
  \bibfield  {author} {\bibinfo {author} {\bibfnamefont {A.}~\bibnamefont
  {Stepanchikova}}, \bibinfo {author} {\bibfnamefont {A.}~\bibnamefont
  {Lagunin}}, \bibinfo {author} {\bibfnamefont {D.}~\bibnamefont {Filimonov}},
  \ and\ \bibinfo {author} {\bibfnamefont {V.}~\bibnamefont {Poroikov}},\
  }\href@noop {} {\bibfield  {journal} {\bibinfo  {journal} {Current medicinal
  chemistry}\ }\textbf {\bibinfo {volume} {10}},\ \bibinfo {pages} {225}
  (\bibinfo {year} {2003})}\BibitemShut {NoStop}%
\bibitem [{\citenamefont {Aderem}(2005)}]{aderem2005systems}%
  \BibitemOpen
  \bibfield  {author} {\bibinfo {author} {\bibfnamefont {A.}~\bibnamefont
  {Aderem}},\ }\href@noop {} {\bibfield  {journal} {\bibinfo  {journal} {Cell}\
  }\textbf {\bibinfo {volume} {121}},\ \bibinfo {pages} {511} (\bibinfo {year}
  {2005})}\BibitemShut {NoStop}%
\bibitem [{\citenamefont {Najafi}\ \emph {et~al.}(2020)\citenamefont {Najafi},
  \citenamefont {Jamali}, \citenamefont {Schober},\ and\ \citenamefont
  {Poor}}]{najafi2020physics}%
  \BibitemOpen
  \bibfield  {author} {\bibinfo {author} {\bibfnamefont {M.}~\bibnamefont
  {Najafi}}, \bibinfo {author} {\bibfnamefont {V.}~\bibnamefont {Jamali}},
  \bibinfo {author} {\bibfnamefont {R.}~\bibnamefont {Schober}}, \ and\
  \bibinfo {author} {\bibfnamefont {H.~V.}\ \bibnamefont {Poor}},\ }\href@noop
  {} {\bibfield  {journal} {\bibinfo  {journal} {IEEE Transactions on
  Communications}\ } (\bibinfo {year} {2020})}\BibitemShut {NoStop}%
\bibitem [{\citenamefont {Weaver}(1991)}]{weaver1991science}%
  \BibitemOpen
  \bibfield  {author} {\bibinfo {author} {\bibfnamefont {W.}~\bibnamefont
  {Weaver}},\ }in\ \href@noop {} {\emph {\bibinfo {booktitle} {Facets of
  systems science}}}\ (\bibinfo  {publisher} {Springer},\ \bibinfo {year}
  {1991})\ pp.\ \bibinfo {pages} {449--456}\BibitemShut {NoStop}%
\bibitem [{\citenamefont {da~F.~Costa}\ \emph {et~al.}(2007)\citenamefont
  {da~F.~Costa}, \citenamefont {Rodrigues}, \citenamefont {Travieso},\ and\
  \citenamefont {Villas~Boas}}]{costa2007characterization}%
  \BibitemOpen
  \bibfield  {author} {\bibinfo {author} {\bibfnamefont {L.}~\bibnamefont
  {da~F.~Costa}}, \bibinfo {author} {\bibfnamefont {F.~A.}\ \bibnamefont
  {Rodrigues}}, \bibinfo {author} {\bibfnamefont {G.}~\bibnamefont {Travieso}},
  \ and\ \bibinfo {author} {\bibfnamefont {P.~R.}\ \bibnamefont
  {Villas~Boas}},\ }\href@noop {} {\bibfield  {journal} {\bibinfo  {journal}
  {Advances in physics}\ }\textbf {\bibinfo {volume} {56}},\ \bibinfo {pages}
  {167} (\bibinfo {year} {2007})}\BibitemShut {NoStop}%
\bibitem [{\citenamefont {Cover}(1999)}]{cover1999elements}%
  \BibitemOpen
  \bibfield  {author} {\bibinfo {author} {\bibfnamefont {T.~M.}\ \bibnamefont
  {Cover}},\ }\href@noop {} {\emph {\bibinfo {title} {Elements of information
  theory}}}\ (\bibinfo  {publisher} {John Wiley \& Sons},\ \bibinfo {year}
  {1999})\BibitemShut {NoStop}%
\bibitem [{\citenamefont {Pincus}(1991)}]{pincus1991approximate}%
  \BibitemOpen
  \bibfield  {author} {\bibinfo {author} {\bibfnamefont {S.~M.}\ \bibnamefont
  {Pincus}},\ }\href@noop {} {\bibfield  {journal} {\bibinfo  {journal}
  {Proceedings of the National Academy of Sciences}\ }\textbf {\bibinfo
  {volume} {88}},\ \bibinfo {pages} {2297} (\bibinfo {year}
  {1991})}\BibitemShut {NoStop}%
\bibitem [{\citenamefont {Fukunaga}(2013)}]{fukunaga2013introduction}%
  \BibitemOpen
  \bibfield  {author} {\bibinfo {author} {\bibfnamefont {K.}~\bibnamefont
  {Fukunaga}},\ }\href@noop {} {\emph {\bibinfo {title} {Introduction to
  statistical pattern recognition}}}\ (\bibinfo  {publisher} {Elsevier},\
  \bibinfo {year} {2013})\BibitemShut {NoStop}%
\bibitem [{\citenamefont {Mullins}(2002)}]{mullins2002database}%
  \BibitemOpen
  \bibfield  {author} {\bibinfo {author} {\bibfnamefont {C.}~\bibnamefont
  {Mullins}},\ }\href@noop {} {\emph {\bibinfo {title} {Database
  administration: the complete guide to practices and procedures}}}\ (\bibinfo
  {publisher} {Addison-Wesley Professional},\ \bibinfo {year}
  {2002})\BibitemShut {NoStop}%
\bibitem [{\citenamefont {Dhar}(2013)}]{dhar2013data}%
  \BibitemOpen
  \bibfield  {author} {\bibinfo {author} {\bibfnamefont {V.}~\bibnamefont
  {Dhar}},\ }\href@noop {} {\bibfield  {journal} {\bibinfo  {journal}
  {Communications of the ACM}\ }\textbf {\bibinfo {volume} {56}},\ \bibinfo
  {pages} {64} (\bibinfo {year} {2013})}\BibitemShut {NoStop}%
\bibitem [{\citenamefont {Hey}\ \emph {et~al.}(2009)\citenamefont {Hey},
  \citenamefont {Tansley},\ and\ \citenamefont {Tolle}}]{hey2009jim}%
  \BibitemOpen
  \bibfield  {author} {\bibinfo {author} {\bibfnamefont {T.}~\bibnamefont
  {Hey}}, \bibinfo {author} {\bibfnamefont {S.}~\bibnamefont {Tansley}}, \ and\
  \bibinfo {author} {\bibfnamefont {K.~M.}\ \bibnamefont {Tolle}},\ }\href@noop
  {} {\enquote {\bibinfo {title} {Jim gray on escience: a transformed
  scientific method.}}\ } (\bibinfo {year} {2009})\BibitemShut {NoStop}%
\bibitem [{Note1()}]{Note1}%
  \BibitemOpen
  \bibinfo {note} {\protect \url {https://www.wikipedia.org/}}\BibitemShut
  {NoStop}%
\bibitem [{\citenamefont {Ratnaparkhi}(1996)}]{ratnaparkhi1996maximum}%
  \BibitemOpen
  \bibfield  {author} {\bibinfo {author} {\bibfnamefont {A.}~\bibnamefont
  {Ratnaparkhi}},\ }in\ \href@noop {} {\emph {\bibinfo {booktitle} {Conference
  on empirical methods in natural language processing}}}\ (\bibinfo {year}
  {1996})\BibitemShut {NoStop}%
\bibitem [{\citenamefont {Sigman}\ and\ \citenamefont
  {Cecchi}(2002)}]{sigman2002global}%
  \BibitemOpen
  \bibfield  {author} {\bibinfo {author} {\bibfnamefont {M.}~\bibnamefont
  {Sigman}}\ and\ \bibinfo {author} {\bibfnamefont {G.~A.}\ \bibnamefont
  {Cecchi}},\ }\href@noop {} {\bibfield  {journal} {\bibinfo  {journal}
  {Proceedings of the National Academy of Sciences}\ }\textbf {\bibinfo
  {volume} {99}},\ \bibinfo {pages} {1742} (\bibinfo {year}
  {2002})}\BibitemShut {NoStop}%
\bibitem [{\citenamefont {Miller}\ \emph {et~al.}(1990)\citenamefont {Miller},
  \citenamefont {Beckwith}, \citenamefont {Fellbaum}, \citenamefont {Gross},\
  and\ \citenamefont {Miller}}]{miller1990introduction}%
  \BibitemOpen
  \bibfield  {author} {\bibinfo {author} {\bibfnamefont {G.~A.}\ \bibnamefont
  {Miller}}, \bibinfo {author} {\bibfnamefont {R.}~\bibnamefont {Beckwith}},
  \bibinfo {author} {\bibfnamefont {C.}~\bibnamefont {Fellbaum}}, \bibinfo
  {author} {\bibfnamefont {D.}~\bibnamefont {Gross}}, \ and\ \bibinfo {author}
  {\bibfnamefont {K.~J.}\ \bibnamefont {Miller}},\ }\href@noop {} {\bibfield
  {journal} {\bibinfo  {journal} {International journal of lexicography}\
  }\textbf {\bibinfo {volume} {3}},\ \bibinfo {pages} {235} (\bibinfo {year}
  {1990})}\BibitemShut {NoStop}%
\end{thebibliography}%

\end{document}